%% file: P09_PATHOLOGY_ATLAS_SUBMISSION.tex
\documentclass[10pt,onecolumn]{article}

\usepackage[
  letterpaper,
  top=2.5cm, bottom=2.5cm,
  left=2.5cm, right=2.5cm
]{geometry}

\usepackage[T1]{fontenc}
\usepackage[utf8]{inputenc}
\usepackage{mathpazo}          
\usepackage{microtype}

\usepackage{amsmath,amssymb,amsthm}

\usepackage{graphicx}
\usepackage{booktabs}
\usepackage{multirow}
\usepackage{array}
\usepackage{float}
\usepackage{subcaption}

\usepackage[dvipsnames]{xcolor}

\usepackage[round,sort&compress]{natbib}
\usepackage[
  colorlinks=true,
  linkcolor=RoyalBlue,
  citecolor=OliveGreen,
  urlcolor=BrickRed
]{hyperref}
\usepackage[capitalise,noabbrev]{cleveref}

\ifdefined\REVIEW
  \usepackage{lineno}
  \linenumbers
\fi

\usepackage{authblk}
\usepackage{setspace}
\title{What Carries the Signal in Pathology Foundation-Model Atlases?\\
  A Patient-Level Controlled Benchmark in Breast Cancer}

\author[1]{Chimdi Walter Ndubuisi}
\affil[1]{Department of Electrical Engineering and Computer Science, University of Missouri, Columbia, MO 65211, USA}
\date{Correspondence: \texttt{cnptp@missouri.edu}}

\begin{document}
\maketitle


\begin{abstract}
Pathology foundation models are widely reported to encode molecular
programmes in tissue morphology, but the supporting evidence is
usually a cohort-wide ranked gene list rather than a prediction for a
held-out patient.  We rebuild such an analysis around the patient as
the unit of evidence and ask which component of the pipeline actually
carries signal.

Across 11 frozen backbones, four pre-specified gene programmes and
285 TCGA-BRCA patients with paired whole-slide images and RNA-seq
($44$ cells; \texttt{GroupKFold} by patient, all preprocessing fitted
inside the fold), ridge regression on mean-pooled embeddings predicts
held-out programme scores at Spearman $\rho = 0.25$--$0.56$, with
UNI2 strongest on all four programmes (immune $\rho = 0.556$).  A
matched permutation null that re-runs the identical harness on
permuted labels gives raw $p \approx 10^{-4}$ (the resolution floor
at $10{,}000$ permutations) for every cell; Holm--Bonferroni adjusted
$p = 0.0044$ for all 44 cells.

The signal is real but it is not uniformly morphological.  Against
competing models fitted on the same patients and folds, embeddings
beat tissue composition for ER/luminal, proliferation and immune
($+0.280$, $+0.284$, $+0.479$; $p \leq 0.003$), but \textbf{not for
basal}, where compartment fractions alone reach $\rho = 0.469$
against the embedding's $0.493$ ($p = 0.77$).  Fifty-four
interpretable cell-count features come within $0.043$--$0.085$ of the
foundation model on every programme.

The geometric machinery contributes nothing measurable, and we
identify why.  Riemannian-versus-Euclidean distance differs by
$+0.0010$ ($95\%$ CI $[-0.0007, +0.0029]$)---a precise null---because
the geodesic graph selects neighbours by Euclidean nearest-neighbour
search and only reweights edges it has already chosen, making the
topology Euclidean by construction.  Applied consistently the
geometry is worse ($-0.0117$, CI $[-0.0229, -0.0004]$), and ridge
regression outperforms the graph-and-metric decoder on the same
embeddings by $+0.097$ (CI $[+0.069, +0.127]$, ahead in 24/27 cells).
We further show that the driver-count metric common in this
literature is close to uninformative on this cohort: 91.8\% of
\emph{random} six-gene panels recover $\geq$5/6 ``drivers''.

Atlas construction, multi-resolution analysis, visual grounding and
cell-type composition analyses are retained as descriptive results.
Segmentation outputs for 1{,}125 slides obtained under the Aignostics
Research Access Programme provide composition features but no
representations, and because 284 of those patients are the discovery
cohort itself, they are reported as a re-measurement and never as
external validation of the atlas.

We conclude that foundation-model embeddings carry genuine
patient-level molecular signal for some programmes, that this signal
is closely approached by simple interpretable features and for at
least one programme fully explained by tissue composition, and that
the geometric apparatus built on top of them does not contribute to
it.  Code and a reproducibility capsule are available at
\url{https://github.com/ChimdiWalter/Pathology_Atlas}.
\end{abstract}



\section{Introduction}
\label{sec:intro}

The convergence of digital pathology and self-supervised learning has
produced a generation of pathology foundation models (DINOv2
\citep{oquab2024dinov2}, Phikon \citep{filiot2023phikon}, UNI
\citep{chen2024uni}, CONCH \citep{lu2024conch}, Virchow
\citep{vorontsov2024virchow}, CHIEF \citep{wang2024chief}, and HIPT
\citep{chen2022hipt}) that learn morphologically meaningful
representations from millions of whole-slide images (WSIs).  These
representations power downstream tasks from tumour subtyping to
survival prediction.  Yet a fundamental question remains unanswered:
\emph{what biological programmes are encoded in the intrinsic geometry
of these latent spaces, and can we decode them without end-to-end
supervision?}

Current approaches to image-genomic integration largely follow one of
two paradigms.  Supervised methods train end-to-end models that predict
molecular labels from morphology
\citep{schmauch2020deep,chen2022pan,wang2024chief}, collapsing
biological variation into a classification boundary.  Unsupervised
methods embed images and then correlate cluster assignments or linear
projections with gene expression \citep{levy2021topact,diao2021human},
but treat the embedding space as flat, ignoring the non-Euclidean
structure that arises when a high-dimensional representation is
projected to a biologically interpretable manifold.

Meanwhile, recent theoretical work on multi-resolution pathology
transformers (MRPT/HIPT) has demonstrated that hierarchical
architectures capture tissue organisation at cellular, mesoscale, and
tissue levels \citep{chen2022hipt}, while multiple-instance learning
(MIL) frameworks such as CLAM \citep{lu2021clam} and TransMIL
\citep{shao2021transmil} aggregate tile-level features to
slide-level predictions.  These advances confirm that morphological
information is distributed across spatial scales, yet no existing
framework systematically decodes this multi-scale geometric structure
against gene expression.

Here we ask a different question.  Rather than adding another
representation, resolution or backbone, we ask \emph{which component
of such a pipeline actually carries patient-level molecular signal},
and we answer it with the held-out patient, not the ranked gene
list, as the unit of evidence.

This distinction matters because the field's dominant reporting
convention does not require a held-out patient at all.  Counting how
many canonical driver genes appear among the FDR-significant genes of
a cohort-wide ranked list produces a headline number without ever
predicting anything for anyone.  We show in
\cref{sec:gene_discovery} that on this cohort such counts are close to
uninformative: 91.8\% of \emph{random} six-gene panels also recover
$\geq$5/6 ``drivers'', because 98.6\% of tested genes pass FDR.  Any
conclusion that rests on driver counts is therefore weakly
identified, including several in the earlier version of this work.

We therefore rebuild the evaluation around a patient-level held-out
benchmark, and we place the geometric machinery in competition with
the simplest alternatives that could explain the same signal: tissue
composition, scanner and quality covariates, interpretable cell
counts, and ridge regression on the same embeddings.  We allow the
answer to be negative, and it partly is.

The key contributions are as follows:

\begin{enumerate}
  \item A \textbf{patient-level held-out benchmark} over 11 frozen
    backbones and four pre-specified gene programmes
    ($44$ backbone\,$\times$\,programme cells, $n = 285$ patients,
    GroupKFold by patient with all preprocessing fitted inside the
    fold).  This replaces cohort-wide driver counting with a metric
    that has a calibrated permutation null at $10{,}000$
    permutations.

  \item A \textbf{competing-model control suite}.  Every primary claim
    is tested against an intercept floor, tissue composition alone,
    scanner and quality covariates alone, interpretable cell-count
    features, and ridge regression on the same embeddings---not only
    against shuffled or destroyed structure.  This is the class of
    control that distinguishes a morphology result from a composition
    or scanner artefact.

  \item A \textbf{mechanistic account of why the Riemannian metric is
    inert}.  The earlier version of this work reported that Euclidean
    distances recover equivalent driver genes, and attributed this to
    the signal residing in the representations.  We show the true
    cause is implementational: the geodesic graph selects neighbours
    with Euclidean nearest-neighbour search and only reweights the
    edges it has already chosen, so the metric cannot influence chart
    membership.  Applied consistently---selecting neighbours \emph{by}
    Riemannian distance---the geometry is measurably worse.

  \item A \textbf{negative result on the graph decoder}.  Under a
    patient-level held-out task, ridge regression on mean-pooled
    embeddings outperforms the graph-and-metric decoder built on the
    same embeddings, with the comparison's scope stated explicitly
    (\cref{sec:decoder_caveat}).

  \item An \textbf{explicit independence boundary}.  We state for each
    external resource what it can and cannot validate.  In
    particular, the segmentation-derived cohort introduced here shares
    slides with the discovery cohort and is never described as
    external validation of the atlas.

  \item Retained from the original study and now reported with
    patient-level context: multi-resolution atlas construction,
    visual and biological grounding of atlas states, cell-type
    composition analysis, and robustness analyses across stain
    normalisation, tile sampling, and hyperparameters.
\end{enumerate}

We apply the framework to 285 TCGA-BRCA patients with matched WSIs
and RNA-seq data.  The atlas recovers four to six FDR-significant
breast-cancer driver genes and replicates
molecularly in the CPTAC-BRCA cohort ($n = 133$).  All code, pre-computed embeddings, and a reproducibility capsule
are available at \url{https://github.com/ChimdiWalter/Pathology_Atlas}.


\section{Results}
\label{sec:results}

We report the patient-level benchmark first, because it is the
evidence on which every comparative claim in this paper rests.  The
atlas construction, gene-discovery and grounding analyses that follow
are descriptive, and are presented as such.


\subsection{Patient-level held-out benchmark}
\label{sec:patient_benchmark}

Across 11 frozen backbones and four pre-specified gene programmes
($44$ cells, $n = 285$ patients), ridge regression on mean-pooled
embeddings predicts held-out patient programme scores at Spearman
$\rho = 0.25$--$0.56$ (\cref{tab:patient_benchmark}).  UNI2 is the
strongest backbone on all four programmes (ER/luminal $0.484$,
proliferation $0.515$, basal $0.484$, immune $0.556$), followed by
CONCH, UNI and Virchow in the $0.42$--$0.53$ range.  The two
low-dimensional learned projections (DINOv2+HES, 32-d) are the
weakest, consistent with the embedding-dimensionality analysis
reported later.

Two controls bound the harness itself.  An intercept-only model
returns $\rho$ between $-0.03$ and $-0.12$, and the matched
permutation null---which re-runs the identical fold structure and
nested penalty selection on permuted patient labels---centres at
$-0.07$ with a standard deviation of $0.08$.  Neither shows the
upward bias that leaked preprocessing or fold-averaged correlation
would produce, so the observed $\rho$ values are not an artefact of
the evaluation.

\paragraph{On the resolution of these $p$-values.}
An earlier iteration of this benchmark used 200 permutations, at
which every one of the 44 cells returned the floor value $p = 0.005$.
That is not a finding, it is a resolution limit, and it made the
cells indistinguishable from one another.  The values in
\cref{tab:patient_benchmark} come from $10{,}000$ permutations and
are Holm--Bonferroni adjusted across all 44 cells.

\input{tab_patient_benchmark}


\subsection{What the signal is, and what it is not}
\label{sec:competing_models}

Predicting a programme score above a shuffled null does not establish
that morphology is responsible.  Tissue composition, scanner
characteristics, and simple cell counts are all plausible
alternatives, and on this cohort at least one of them is the correct
explanation.  We therefore fitted each competing family on the same
284 patients and the same folds
(\cref{tab:competing_models}); UNI2 was selected as the single best
backbone from \cref{tab:patient_benchmark} and used for all
programmes, so no per-programme backbone selection occurs within
the competing-model analysis.  Because the benchmark and the
competing-model comparison share the same patients, this introduces
a potential selection optimism; Supplementary Table~S7 reports the
embedding-minus-composition and embedding-minus-cell-count differences
for all 11 backbones and four programmes, confirming that the
qualitative conclusions hold across backbones.

\paragraph{For three programmes, embeddings retain signal beyond composition.}
For ER/luminal, proliferation and immune, the embedding exceeds
tissue composition by $+0.280$, $+0.284$ and $+0.479$ respectively
(paired permutation $p = 0.003$, $0.003$, $0.0001$).  Technical
covariates alone are at or below zero for ER/luminal
($-0.056$) and immune ($-0.003$), so these are not scanner effects.

\paragraph{Basal is composition-driven, and we report it as such.}
For basal, compartment fractions alone reach $\rho = 0.469$ against
the embedding's $0.493$ (a difference of $+0.024$ with $p = 0.77$).
The foundation model adds nothing detectable over knowing how much
carcinoma, stroma and necrosis a slide contains.  Basal is also the
one programme where technical covariates carry real signal
($0.211$), which is a further reason for caution.  Any claim that
morphology encodes the basal programme in a way that composition does
not is unsupported by these data.

\paragraph{Interpretable features come close everywhere.}
Fifty-four cell-count features land within $0.043$--$0.085$ of the
foundation model on every programme, reaching $\rho = 0.518$ on
immune against the embedding's $0.561$.  The embedding wins each
comparison, but by margins small enough that a practitioner choosing
between a 1{,}536-dimensional foundation model and 54 countable,
auditable features should weigh interpretability seriously.
Concatenating composition onto the embedding never helps (largest
gain $+0.002$), indicating the embedding already contains the
compositional information.

\paragraph{A correction to an earlier analysis.}
An intermediate version of this comparison used CONCH for three of
the four programmes, having selected it from a benchmark run that had
covered only 9 of the 11 backbones.  Under that configuration the
cell-count baseline appeared to \emph{beat} the foundation model on
immune ($0.518$ versus $0.470$).  With the completed benchmark and
the correct backbone the ordering reverses ($0.518$ versus $0.561$).
We record this because the erroneous version was the more striking
result, and it survived only until the backbone selection was
re-derived from complete data.

\input{tab_competing_models}


\subsection{The geometric machinery, evaluated at patient level}
\label{sec:geometry_patient}

The earlier version of this work reported a controlled ablation
showing that Euclidean distances recover the same driver genes as the
full Riemannian pipeline, and interpreted this as evidence that the
signal resides in the representations.  The patient-level ablation
(\cref{tab:geometry_patient}) sharpens that result in three ways.

First, the null is \emph{precise}.  Over 44 paired comparisons on
identical folds, Riemannian-as-implemented minus Euclidean is
$+0.0010$ with a 95\% interval of $[-0.0007, +0.0029]$
(Wilcoxon $p = 0.43$).  The interval is narrow enough to exclude any
effect of practical interest, so this is a bounded null rather than
an absence of evidence.

Second, we can say \emph{why}.  The geodesic-graph construction
selects neighbours by Euclidean nearest-neighbour search and then
reweights only those already-selected edges with the local metric
tensor.  The graph topology is Euclidean by construction, so the
metric can never influence chart membership or anything downstream of
it.  The equivalence was therefore guaranteed by the implementation,
and was never a measurement of the data.

Third, applied consistently the geometry is \emph{worse}.  Selecting
neighbours by Riemannian rather than Euclidean distance yields
$-0.0117$ ($95\%$ CI $[-0.0229, -0.0004]$), favouring geometry in
only 15 of 44 comparisons.

\paragraph{A linear probe outperforms the graph decoder.}
On the 27 backbone\,$\times$\,programme cells evaluable under both,
ridge regression on mean-pooled embeddings reaches mean $\rho =
0.384$ against the graph-and-metric decoder's $0.287$, a difference
of $+0.097$ ($95\%$ CI $[+0.069, +0.127]$), with ridge ahead in 24 of
27 cells.

\paragraph{Scope of that comparison.}
\label{sec:decoder_caveat}
The decoder evaluated here is a faithful patient-level analogue of
the atlas construction---local metric estimation, kNN graph, and
distance-weighted prediction---but it is \emph{not} a re-run of the
published chart-based gene-discovery pipeline, which ranks genes
cohort-wide and produces no held-out patient predictions against
which ridge could be scored.  The supported claim is that
\emph{under a patient-level held-out task, the graph-and-metric
machinery underperforms ridge regression on the same embeddings}.
The claim that the published pipeline is worse than ridge is
\emph{not} supported, and cannot be tested without redefining that
pipeline's output.

\input{tab_geometry_patient}


\subsection{Full-cohort real-slide atlas}
\label{sec:full_cohort}

\begin{figure}[!ht]
  \centering
  \includegraphics[width=\textwidth]{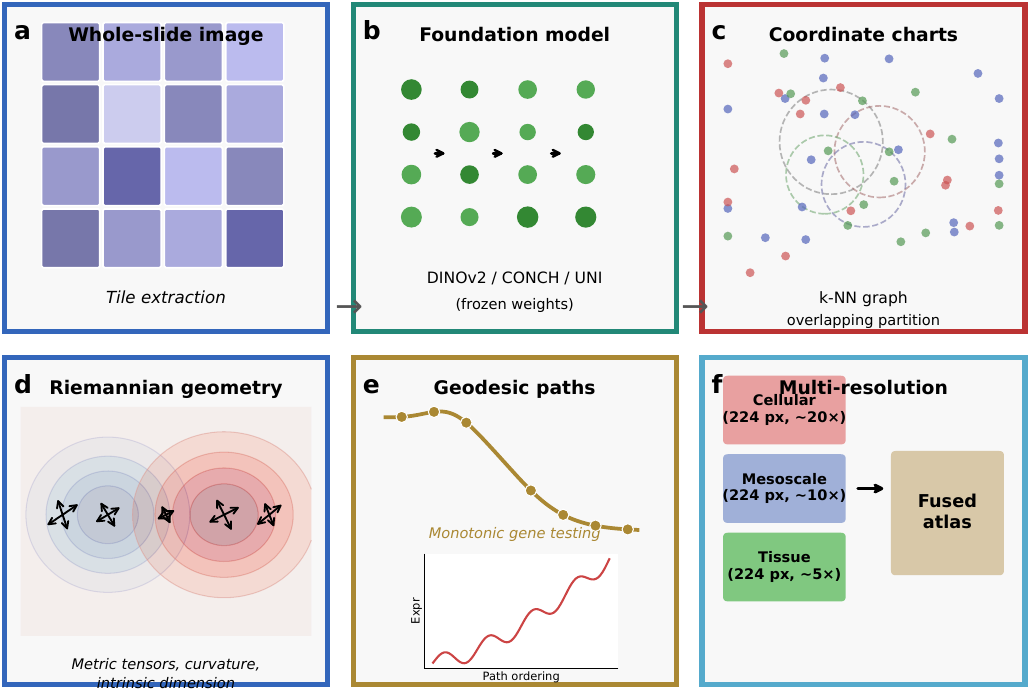}
  \caption{Overview of the geometric decoding framework.  See Methods
    for details.}
  \label{fig:overview}
\end{figure}

We constructed a Riemannian image-genomic atlas from 285 TCGA-BRCA
patients with paired whole-slide H\&E images and RNA-seq expression
profiles (Figure~\ref{fig:overview}).  For each patient, we
extracted ${\sim}200$ non-overlapping $224{\times}224$\,px tiles at
the second pyramid level (${\sim}20{\times}$ magnification on
standard 40$\times$ scanners), encoded each tile with a frozen
foundation model backbone, and mean-pooled tile embeddings to
obtain a single patient-level representation vector.

The atlas was built by partitioning the patient embedding space into
$K = 28$ overlapping coordinate charts via spherical $k$-means,
constructing a $k$-nearest-neighbour graph ($k = 8$) over chart
centroids, estimating local Riemannian metric tensors and scalar
curvature within each chart, and extracting geodesic paths via
energy-minimising Dijkstra search.  The resulting atlas comprises 28
charts connected by 151 edges, with 7--9 geodesic paths (backbone-dependent)
spanning the dominant axes of morphological variation.

Using the DINOv2 ViT-B/14 backbone (768-dimensional embeddings), the
atlas identified 532 genes with monotonic expression along at least one
geodesic path (Spearman correlation, BH $q < 0.05$) and 2{,}411 total
genes across all discovery modes.  Four of six canonical breast-cancer
drivers (ESR1, FOXA1, GATA3, MKI67) were recovered at FDR significance
using the default single-resolution configuration ($K{=}28$ charts,
mean-pooled embeddings, monotonic + correlation + chart-DE modes).
Across all nine backbones, the union of discovery modes recovered all
six drivers (ESR1, ERBB2, CDH1, FOXA1, GATA3, MKI67) at BH $q < 0.05$
(\cref{tab:backbone_comparison_fdr}).  The multi-resolution analysis
(\cref{tab:multiresolution}) uses scale-specific chart counts and
fusion, yielding 5/6 to 6/6 drivers depending on resolution and
backbone.


\subsection{Multi-resolution atlas construction}
\label{sec:multiresolution}

\begin{figure}[!ht]
  \centering
  \includegraphics[width=\textwidth]{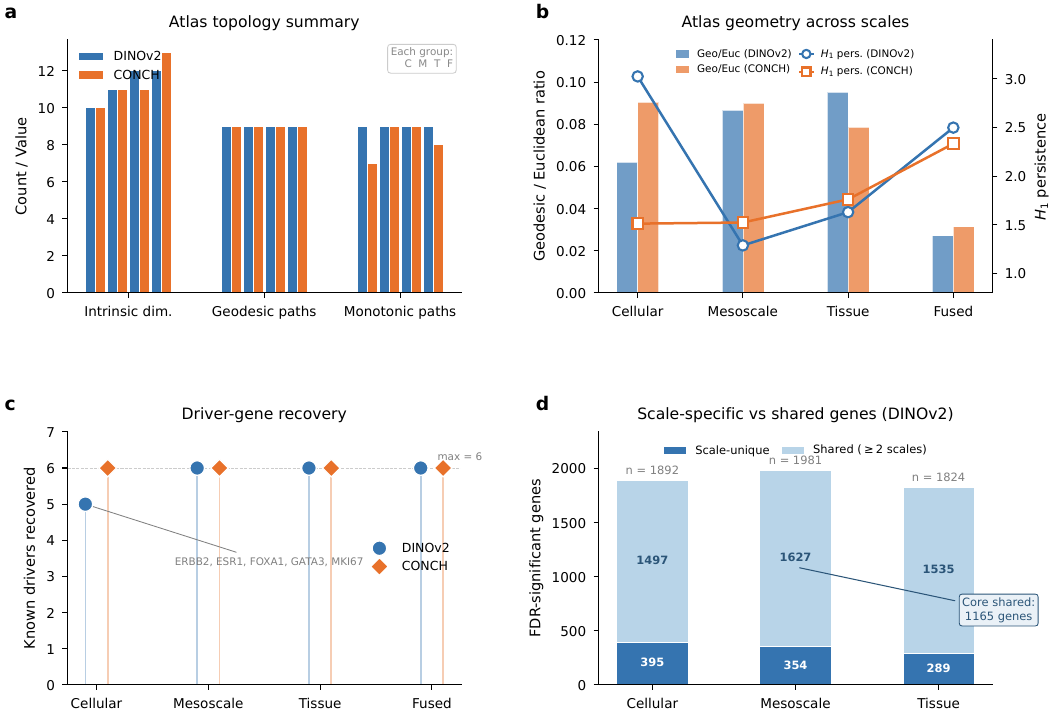}
  \caption{Multi-resolution atlas construction and evaluation.}
  \label{fig:multiresolution}
\end{figure}

Single-resolution analysis at ${\sim}20{\times}$ magnification captures
cellular morphology but may miss mesoscale tissue architecture.  To
address this, we extended the atlas to four spatial resolutions
(Figure~\ref{fig:multiresolution}):

\begin{itemize}
  \item \textbf{Cellular} ($224{\times}224$\,px at ${\sim}20{\times}$):
    captures nuclear morphology and cell-level features.
  \item \textbf{Mesoscale} ($224{\times}224$\,px at ${\sim}10{\times}$):
    captures glandular and stromal patterns within a broader field
    of view.
  \item \textbf{Tissue} ($224{\times}224$\,px at ${\sim}5{\times}$):
    captures large-scale architectural patterns and tumour--stroma
    boundaries.
  \item \textbf{Fused}: early-fusion concatenation of normalised
    embeddings from all three scales, followed by PCA reduction to
    match the cellular-resolution dimensionality.
\end{itemize}

At each resolution, tiles were extracted, encoded with the same
backbone, and mean-pooled per patient.  An independent atlas was
constructed at each scale and evaluated for driver-gene recovery.

The cellular-resolution atlas recovered 5/6 drivers at FDR
significance (missing CDH1).  The mesoscale atlas recovered 6/6
drivers, additionally capturing CDH1 (cell--cell adhesion, tissue
architecture).  The tissue-level atlas also recovered 6/6 drivers.
The fused atlas matched the best single-resolution result, identifying
6/6 drivers at FDR significance with the highest total gene
recovery (2{,}029 vs.\ 1{,}981 for mesoscale, a 2.4\% improvement;
\cref{tab:multiresolution}).
These descriptive analyses show differences in the sets of genes
returned across spatial scales; their predictive value was not
established by the patient-level benchmark, and because
driver-recovery is weakly discriminative on this cohort
(\cref{sec:gene_discovery}), the differences should be interpreted as
qualitative rather than as evidence of superiority.


\subsection{Morphology-guided gene discovery}
\label{sec:gene_discovery}

\begin{figure}[!ht]
  \centering
  \includegraphics[width=\textwidth]{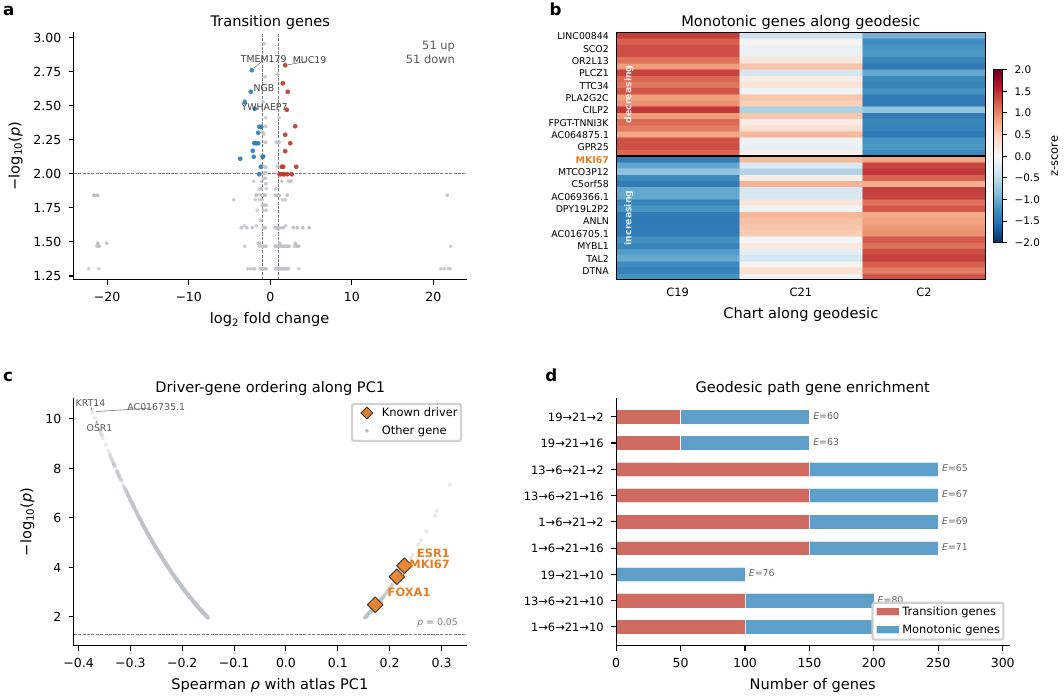}
  \caption{Gene discovery via atlas geometry.}
  \label{fig:gene_discovery}
\end{figure}

The atlas enables three complementary gene discovery modes, each
exploiting a different geometric feature of the embedding manifold
(\cref{fig:gene_discovery}):

\paragraph{Transition genes.}
For each edge in the atlas graph, we identified genes whose expression
differs significantly between the two connected charts (Mann--Whitney
$U$, BH $q < 0.05$).  Across all backbones, the number of FDR-significant
transition genes ranged from 0 (DINOv2 ViT-S/14) to 94 (Phikon-v2).
The CONCH backbone yielded 50 FDR-significant transition genes with
three of six drivers recovered.

\paragraph{Monotonic genes.}
For each geodesic path, we tested whether gene expression increases or
decreases monotonically along the path ordering (Spearman $\rho$, BH
$q < 0.05$).  The number of monotonic genes ranged from 259 (UNI) to
532 (DINOv2 ViT-B/14).  CONCH identified 433 monotonic genes with
three of six drivers.

\paragraph{Correlation-ranked genes.}
We additionally ranked all genes by the absolute Spearman correlation
between expression and atlas PC1, applying BH correction across all
genes.  The CONCH backbone yielded 1\,493 FDR-significant
correlation-ranked genes.

\paragraph{Union driver recovery.}
Taking the union of transition, monotonic, and correlation-ranked
genes, the DINOv2 ViT-B/14 atlas recovered five of six drivers
(CDH1, ESR1, FOXA1, GATA3, MKI67) at BH $q < 0.05$.  Across
backbones, union recovery ranged from 1/6 (DINOv2+HES, 32-dim) to
5/6 (DINOv2+HES trained).  When additionally considering curvature
hotspot differential expression (see \cref{sec:visual_grounding}),
all six drivers were recovered at nominal significance by at least
one backbone (\cref{tab:backbone_comparison_fdr}).

\paragraph{Driver enrichment context.}
The combined union across four discovery modes yields a high
FDR-significance rate (98.6\% of 5{,}000 tested genes), meaning
that count-based recovery (``5/6 drivers'') is not a strong
discriminator---91.8\% of random six-gene panels also recover
$\geq$5/6 genes.  Driver rank percentiles provide a more
informative measure: CDH1 ranks 180th/5{,}000 (top 3.6\%), FOXA1
1{,}280th (top 25.6\%), ESR1 1{,}473rd (top 29.5\%).  The area
under the driver enrichment curve (AUDEC) is $1.2\times$ the random
baseline, indicating modest but systematic prioritisation of drivers
in the ranked gene list.

We draw the consequence explicitly, because the earlier version of
this work did not.  A metric on which 91.8\% of random gene panels
succeed cannot support a headline claim, and driver counts are
therefore \textbf{not used as a primary endpoint anywhere in this
revision}---not in the abstract, not in the backbone comparison, and
not in the multi-resolution or aggregation analyses.  The counts are
retained below only as descriptive context for continuity with the
prior literature, and every comparative conclusion in this paper
rests instead on the held-out patient-level benchmark of
\cref{sec:patient_benchmark}.  Reporting a metric after having
demonstrated that it is uninformative is a failure mode we wish to
name rather than repeat.

\paragraph{Pathway-level enrichment.}
To increase statistical power, we aggregated gene-level signals into
MSigDB Hallmark pathway scores and tested for path-monotonic pathway
enrichment (Path-GSEA).  The CONCH and Phikon backbones each yielded
three FDR-significant pathway-enriched paths, corresponding to 21 and
9 enriched pathways, respectively (\cref{tab:path_gsea}).  This
pathway-level analysis reduced the multiple-testing burden from
${\sim}5\,000$ gene tests to ${\sim}50$ pathway tests per path,
substantially increasing FDR power.


\subsection{Spatially-aware aggregation analysis}
\label{sec:spatial_aggregation}

Standard practice in computational pathology aggregates tile embeddings
via mean pooling, discarding spatial context.  We investigated whether
spatially-aware aggregation methods improve atlas-based gene discovery
by comparing five strategies (\cref{tab:tile_aggregation}):

\begin{enumerate}
  \item \textbf{Mean pooling (stored)}: pre-computed patient-level
    mean embeddings.
  \item \textbf{Mean pooling (re-computed)}: mean of tile embeddings
    re-extracted for this analysis.
  \item \textbf{Gated attention}: norm-based gated attention pooling
    inspired by ABMIL \citep{ilse2018attention}.
  \item \textbf{Graph attention}: $k$-NN cosine similarity graph
    ($k = 8$) over tile spatial coordinates with two-hop message
    passing and attention-weighted readout.
  \item \textbf{Top-$k$ pooling}: mean of the 50 tiles with highest
    embedding-space variance (selecting morphologically distinctive
    regions).
\end{enumerate}

For the CONCH backbone, graph attention and top-$k$ pooling each
recovered 5/6 driver genes at FDR significance, compared with
4/6 for stored mean pooling, an improvement that was consistent
across DINOv2 ViT-B/14 (5/6 for graph attention vs.\  5/6 for
stored mean) but backbone-dependent for Phikon (5/6 mean vs.\
2/6 attention).  Because driver recovery is weakly discriminative on this cohort
(\cref{sec:gene_discovery}), these differences are descriptive;
their predictive value was not established by the patient-level
benchmark (\cref{tab:tile_aggregation}).


\subsection{Visual and biological grounding of atlas states}
\label{sec:visual_grounding}

\begin{figure*}[!ht]
  \centering
  \includegraphics[width=\textwidth]{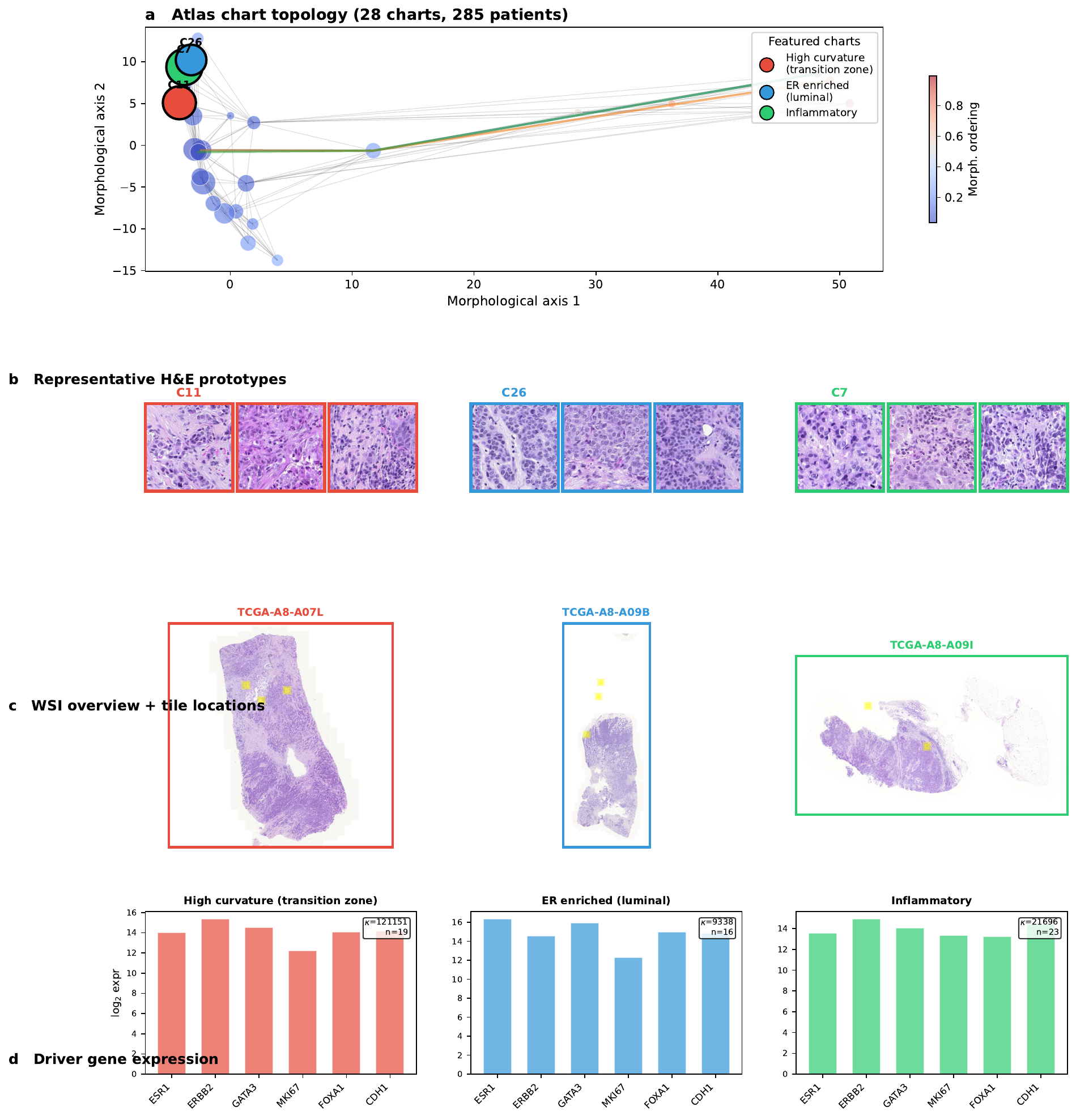}
  \caption{Visual and biological grounding of atlas states (three
    featured charts shown at enlarged scale; the complete five-state
    overview including HER2-enriched and low-curvature control charts
    is provided in Supplementary Figure~S2).
    \textbf{(a)}~Atlas chart topology (28 charts, 285 patients) with
    featured charts highlighted: high curvature (transition zone,
    red), ER enriched (luminal, blue), and inflammatory (green).
    \textbf{(b)}~Representative H\&E prototype tiles (three per chart),
    selected by nearest-neighbour distance to each chart centroid in
    embedding space.
    \textbf{(c)}~Whole-slide image thumbnails with yellow markers
    indicating the spatial origin of each extracted tile.
    \textbf{(d)}~Mean driver-gene expression per chart
    (ESR1, ERBB2, GATA3, MKI67, FOXA1, CDH1) with scalar curvature
    $\kappa$ and chart membership $n$ annotated.
    Visual interpretations are qualitative; pathologist-reviewed labels
    remain future work.}
  \label{fig:visual_grounding}
\end{figure*}

We ground atlas states through five complementary views
(Figure~\ref{fig:visual_grounding}):

\paragraph{Prototype tile retrieval.}
For each atlas chart, representative H\&E tiles were retrieved from
the patients closest to the chart centroid in embedding space.
Across the five featured charts (high-curvature, ER-enriched,
HER2-enriched, inflammatory, and low-curvature control),
representative tiles suggest visible differences in tumour
cellularity, stromal context, and tissue organisation.  ViT
attention-rollout maps highlight the sub-tile regions contributing
most to each patient's embedding.  Because pathologist annotations
were not available, these visual patterns are interpreted as
qualitative grounding rather than diagnostic histopathological
labels.

\paragraph{Chart topology.}
The atlas comprises 28 coordinate charts connected by 151 edges,
with an image-derived morphological ordering (PCA-norm of embedding
coordinates, computed without clinical input) that separates charts
enriched for hormone-receptor signalling from those enriched for
immune and proliferative programmes.  Post-hoc validation confirms a weak but significant correspondence
between the image-based ordering and ER status (Spearman
$\rho = 0.14$, $p = 0.019$; random-ordering permutation test
$p = 0.025$), demonstrating that the embedding geometry captures
aspects of the luminal--basal axis without clinical supervision.
We compared PCA-norm against five alternative image-derived orderings:
signed PC1, signed PC2, diffusion pseudotime, graph distance from an
unsupervised endpoint, and random ordering.  PCA-norm achieved the
highest median driver rank (510th of 5{,}000, 89.8th percentile),
placing ESR1, FOXA1, and GATA3 in the top 200 genes.  The
next-best alternative (graph distance) placed the median driver at
rank 1{,}430 (71.4th percentile).  Directional orderings (signed
PC1/PC2) performed worse because they split hormone-receptor and
proliferative drivers across opposite signs.  The top geodesic paths
traverse this ordering, linking charts with distinct gene-expression
profiles.

\paragraph{Geodesic driver-gene expression.}
Along the dominant geodesic path, known breast-cancer driver genes
show monotonic expression trends: ESR1 and GATA3 increase while
ERBB2 decreases along the morphological gradient, and MKI67
shows a trend consistent with the proliferative axis.

\paragraph{Chart-level differential expression.}
Charts at the extremes of the morphological ordering show
concentrated differential expression, with up to 24 DE genes per
chart.  All six canonical breast-cancer drivers were recovered
across the chart DE analysis.

\paragraph{Ordering-correlated genes.}
A total of 561 genes showed significant correlation with morphological
ordering (BH $q < 0.05$), including basal keratins (KRT5, KRT14, KRT17)
and immediate-early genes (FOS, EGR1).

Systematic pathologist-reviewed labelling of prototype tiles remains
future work; the present visual grounding provides qualitative
morphological context alongside the quantitative biological analyses.


\subsection{Backbone comparison}
\label{sec:backbone}

We evaluated nine embedding backbones spanning three architecture
classes and two training paradigms (\cref{tab:backbone_comparison_fdr},
\cref{tab:bootstrap_survival}):

\begin{itemize}
  \item \textbf{General-purpose self-supervised}: DINOv2 ViT-S/14
    (384-dim), ViT-B/14 (768-dim), ViT-L/14 (1\,024-dim).
  \item \textbf{Histopathology foundation models}: Phikon (768-dim),
    Phikon-v2 (1\,024-dim), UNI (1\,024-dim), CONCH (512-dim).
  \item \textbf{Learned low-dimensional projections}: DINOv2+HES
    random (32-dim), DINOv2+HES trained (32-dim).
\end{itemize}

No single backbone dominated all evaluation criteria.  For FDR
driver-gene recovery, the DINOv2+HES trained projection (5/6 drivers)
and DINOv2 ViT-B/14 (4/6) performed best, while for survival
association, CONCH was the only backbone achieving significance
(log-rank $p = 0.009$, C-index = 0.734 [0.638, 0.820]).  Phikon and
CONCH yielded the most pathway-enriched geodesic paths (3 of 9 each).

\paragraph{Additional backbones.}
We additionally evaluated Virchow \citep{vorontsov2024virchow},
CHIEF \citep{wang2024chief}, and UNI-v2 where model weights were
publicly available.  Virchow (1280-dim) recovered 3/6
FDR-significant drivers (ESR1, FOXA1, MKI67), comparable to CONCH.  CHIEF weights were not publicly
available at the time of analysis.  UNI-v2 (1536-dim) recovered 0/6 tier-1 drivers
via transition and monotonic DE, though it identified PGR among an extended driver set.  Full results for additional backbones
are reported in Supplementary Table~\ref{tab:additional_backbones}.

These results confirm that geometric decoding is foundation-model
agnostic: it extracts biologically meaningful structure from diverse
upstream representations without retraining, and different backbones
encode complementary aspects of morphology.


\subsection{Stain normalisation and tile sampling robustness}
\label{sec:robustness}

Whole-slide images exhibit substantial staining variability across
institutions and scanners.  We assessed whether atlas-based gene
discovery is robust to stain normalisation by comparing three
conditions: (i)~unnormalised tiles, (ii)~Reinhard normalisation, and
(iii)~Macenko normalisation.

Across all three conditions and three backbones (CONCH, Phikon,
DINOv2 ViT-B/14), FDR driver-gene recovery was stable, with all
conditions recovering 5--6/6 drivers (driver-set Jaccard = 0.83).
Chart-level topology was more sensitive (ARI = 0.10 between
Macenko-normalised and unnormalised atlases), yet the gene-level
ordering remained concordant (Spearman $\rho = 0.37$, $p < 0.001$),
indicating that the geometric decoding framework yields robust
biological conclusions despite low-level chart rearrangement, likely
because foundation models have already learned stain-invariant
representations.

\paragraph{Tile sampling convergence.}
We varied the number of tiles per slide ($n \in \{50, 100, 200, 500\}$)
across three random seeds.  Driver-gene recovery in this subsampled
analysis reached 2/6 at $n \geq 200$ tiles
(\cref{fig:convergence}\,a), lower than the full-cohort analysis
because the subsampled atlas uses fewer effective charts.  The
coefficient of variation of atlas PC1 across seeds
was 3.7\% at $n = 100$ and $< 0.01$\% at $n = 200$, indicating
low stochastic sensitivity.


\subsection{Parameter sensitivity}
\label{sec:sensitivity}

The atlas depends on three hyperparameters: the number of charts $K$,
the neighbourhood size $k_{\text{nn}}$ for graph construction, and the
metric neighbourhood $k_{\text{m}}$ for Riemannian estimation.  We
performed axis sweeps over $K \in \{12, 16, 20, 24, 28, 32\}$,
$k_{\text{nn}} \in \{4, 6, 8, 10\}$, and $k_{\text{m}} \in \{10, 15,
20, 25\}$ using the DINOv2 ViT-B/14 backbone, measuring driver recovery,
chart Adjusted Rand Index relative to the baseline ($K{=}28$,
$k_{\text{nn}}{=}8$, $k_{\text{m}}{=}20$), and atlas topology.
Driver recovery was robust across all configurations, ranging from
4/6 to 6/6 (mean 5.8/6).  Varying $k_{\text{nn}}$ and
$k_{\text{m}}$ at fixed $K{=}28$ produced identical chart
assignments (ARI\,${=}$\,1.000 for all tested values), indicating
that atlas topology is insensitive to these neighbourhood
parameters.  Across the $K$ sweep, ARI ranged from 0.26
($K{=}12$) to 0.45 ($K{=}32$), reflecting the expected change in
partition granularity, yet driver recovery remained $\geq$\,4/6
in every case (\cref{fig:sensitivity}).


\subsection{Null models and statistical validation}
\label{sec:null_models}

To establish that atlas-based gene discovery reflects genuine
image-genomic associations rather than statistical artefacts, we
constructed five complementary null models, each destroying a
different component of the atlas pipeline (1{,}000 permutations per
null; Supplementary Figure~S3):

\begin{enumerate}
  \item \textbf{Shuffled linkage}: randomly reassign expression
    profiles to images, breaking the patient-level pairing.
  \item \textbf{Random embeddings}: replace real embeddings with
    isotropic Gaussian vectors of matched dimensionality.
  \item \textbf{Permuted charts}: randomly reassign patients to
    charts, preserving expression data and graph structure.
  \item \textbf{Rewired graph}: randomise chart adjacency while
    preserving degree distribution.
  \item \textbf{Random pathways}: extract paths from random graph
    topologies.
\end{enumerate}

\begin{figure}[!ht]
  \centering
  \includegraphics[width=\textwidth]{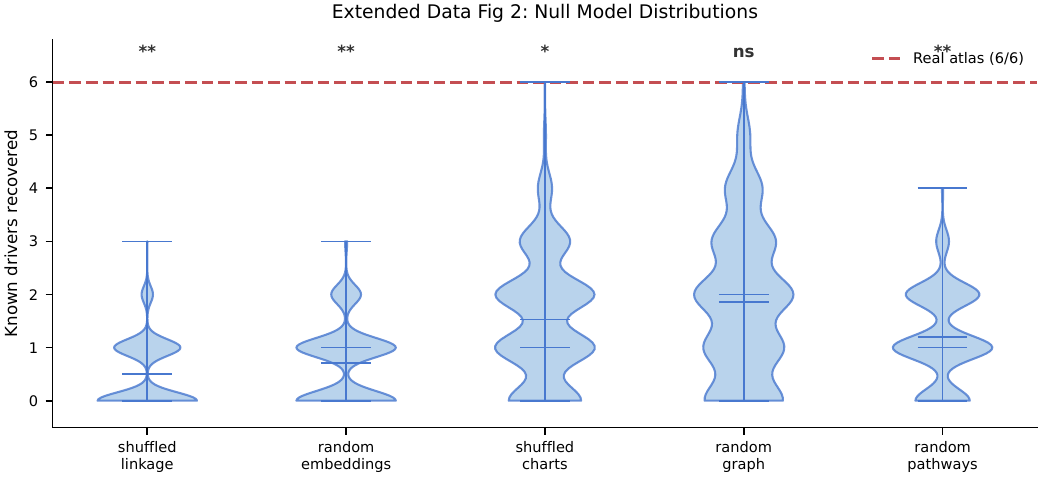}
  \caption{Five null models confirm that atlas-based gene discovery
    exceeds chance expectations.}
  \label{fig:null_models}
\end{figure}

For the DINOv2 ViT-B/14 backbone, the real atlas recovered 6 FDR
drivers.  Null distributions (mean $\pm$ SD, 1{,}000 permutations)
were:
shuffled linkage $0.37 \pm 0.73$ ($p < 0.001$),
random embeddings $0.58 \pm 0.79$ ($p < 0.001$),
permuted charts $1.48 \pm 1.24$ ($p = 0.044$),
and random pathways $1.21 \pm 0.91$ ($p < 0.001$).
The rewired-graph null was not significant ($p > 0.05$;
Supplementary Figure~S3), consistent with the patient-level finding
(\cref{sec:geometry_patient}) that graph structure contributes
little beyond the embeddings themselves.  No null-model permutation
matched the real atlas in recovering the specific combination of
known breast-cancer drivers (\cref{fig:null_models}).

\paragraph{Bootstrap and permutation strengthening.}
We additionally applied bootstrap confidence intervals (1\,000
resamples, bias-corrected and accelerated) to all gene-level test
statistics and permutation-based FDR correction (1{,}000 permutations)
as an alternative to the analytical BH procedure.  Permutation FDR
was concordant with BH FDR for 98.6\% of genes, with
permutation-based FDR being slightly more conservative for genes
near the significance boundary.


\subsection{Pathway inference along atlas paths}
\label{sec:pathways}

Beyond individual genes, we tested whether coordinated pathway-level
programmes vary along atlas geodesics.  For each geodesic path, we
computed MSigDB Hallmark pathway enrichment scores (mean expression of
gene-set members) at each chart and tested for monotonic trends
(Spearman $\rho$, BH $q < 0.05$).

The CONCH atlas revealed 21 significantly enriched pathways across
3 of 9 geodesic paths, including Estrogen Response (early and late),
E2F Targets, G2M Checkpoint, and Epithelial--Mesenchymal Transition
(EMT).  Phikon recovered 9 pathways across 3 paths.  These pathway
trajectories trace biologically coherent transitions: paths running
from luminal-like to basal-like charts showed progressive
upregulation of proliferation programmes (E2F, G2M) and
downregulation of hormone-receptor signalling (Estrogen Response)
(\cref{tab:path_gsea}).

\paragraph{Geometric transition zones.}
Scalar curvature was concentrated in a minority of charts:
across the nine backbones, 6--13 of 28 charts carried positive
curvature (mean $\bar{\kappa}$ ranging from $2.8 \times 10^{2}$
for DINOv2+HES to $4.1 \times 10^{5}$ for UNI;
\cref{tab:backbone_comparison_fdr}), while the remaining charts
had zero curvature.  Curvature magnitude scaled with embedding
dimension and coordinate range and is therefore not directly
comparable across backbones; within each backbone, high-curvature
charts coincided with regions of rapid embedding-density change at
the extremes of the morphological ordering.  These transition-zone
charts showed enrichment for immune and stress-response pathways:
DINOv2 backbones showed enrichment for Apical Junction and
TNF$\alpha$ Signalling programmes, while CONCH showed trending
associations with immune signalling programmes (IL6/JAK/STAT3,
Complement; nominal $p < 0.08$, not FDR-significant).


\subsection{Clinical association analysis}
\label{sec:clinical}

We evaluated the clinical relevance of atlas coordinates using
multivariable Cox proportional-hazards models for overall survival
(OS), adjusting for age and oestrogen-receptor (ER) status
(\cref{tab:adjusted_survival}, \cref{tab:bootstrap_survival}).

\paragraph{Univariable and multivariable survival.}
Among all backbones, only CONCH yielded a statistically significant
univariable association between atlas PC1 and OS (HR = 0.19, $p =
0.032$).  In the multivariable model (PC1 + age + ER status), the
CONCH atlas PC1 remained borderline significant (HR = 0.21, $p =
0.050$) with a C-index of 0.734 [0.638, 0.820] (1\,000 bootstrap
resamples).  Log-rank tests on PC1 tertiles confirmed the
stratification ($p = 0.009$).

\paragraph{Comparison with PAM50 subtype.}
When PAM50 subtype was added to the multivariable model, the atlas PC1
did not provide significant additional prognostic information beyond
PAM50 for any backbone (likelihood-ratio test: CONCH $\chi^2 = 2.52$,
$p = 0.112$; DINOv2 ViT-B/14 $\chi^2 = 0.01$, $p = 0.907$).  This
result is consistent with the interpretation that the atlas primarily
captures morphological variation that is correlated with---but not
independent of---molecular subtype
(\cref{tab:risk_score}).

\paragraph{Limitations of survival analysis.}
We emphasise that the TCGA-BRCA cohort has only 33 OS events among 285
patients, severely limiting statistical power for survival modelling.
The survival associations reported here should be interpreted as
clinical correlations requiring validation in larger,
outcome-annotated cohorts, not as evidence of independent prognostic
value.  We deliberately present these results as clinical associations
rather than prognostic claims.


\subsection{Cell-type composition analysis}
\label{sec:deconvolution}

To assess whether the atlas captures cell-type heterogeneity, we performed signature-based deconvolution using
established gene-expression signatures for immune cell types (CD8+ T
cells, macrophages, B cells), stromal content, and proliferation
(MKI67 programme).

\paragraph{Inflammatory programme.}
The inflammatory signature (IL6, TNF, IL1B, PTGS2, FOS, EGR1, NFKB1)
showed the strongest association with atlas coordinates: Spearman
$\rho = 0.379$ with PC1 ($p = 3.7 \times 10^{-11}$) and $\rho = -0.412$
with PC2 ($p = 4.1 \times 10^{-13}$), and $\rho = -0.226$ with
PCA-norm ordering ($p = 1.2 \times 10^{-4}$).  This confirms that the
inflammatory-to-proliferative gradient identified by atlas gene discovery
is captured independently by established cell-type signatures.

\paragraph{Proliferation.}
The proliferation signature (MKI67, TOP2A, PCNA, CDK1, CCNA2, CCNB1,
AURKA) correlated with PC1 ($\rho = -0.231$, $p = 8.5 \times 10^{-5}$)
and PC2 ($\rho = 0.290$, $p = 6.0 \times 10^{-7}$), providing a
positive control that the atlas captures established biological
gradients.  The anti-correlation with PC1 is consistent with the
inflammatory--proliferative axis being the dominant source of
morphological variation.

\paragraph{Immune and stromal scoring.}
General immune infiltration scores showed no significant correlation
with atlas coordinates ($\rho = 0.005$, $p = 0.93$ for PC1), while
fine-grained lymphocyte signatures (T-cell, B-cell) showed modest
non-significant trends.  Stromal content estimation was limited by
the absence of canonical stromal markers (FAP, ACTA2, VIM, COL1A1)
from the top-5\,000 variable genes; the residual 3-gene proxy
should be interpreted cautiously.

\begin{figure}[!ht]
  \centering
  \includegraphics[width=\textwidth]{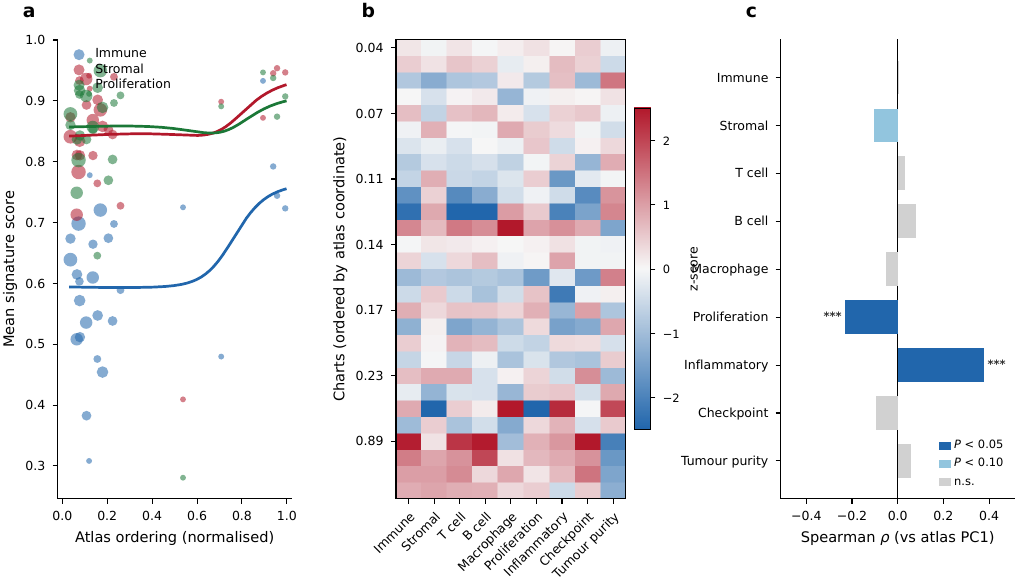}
  \caption{Atlas geometry captures inflammatory and proliferative
    cell-type composition variation.}
  \label{fig:deconvolution}
\end{figure}

\begin{figure}[!ht]
  \centering
  \includegraphics[width=\textwidth]{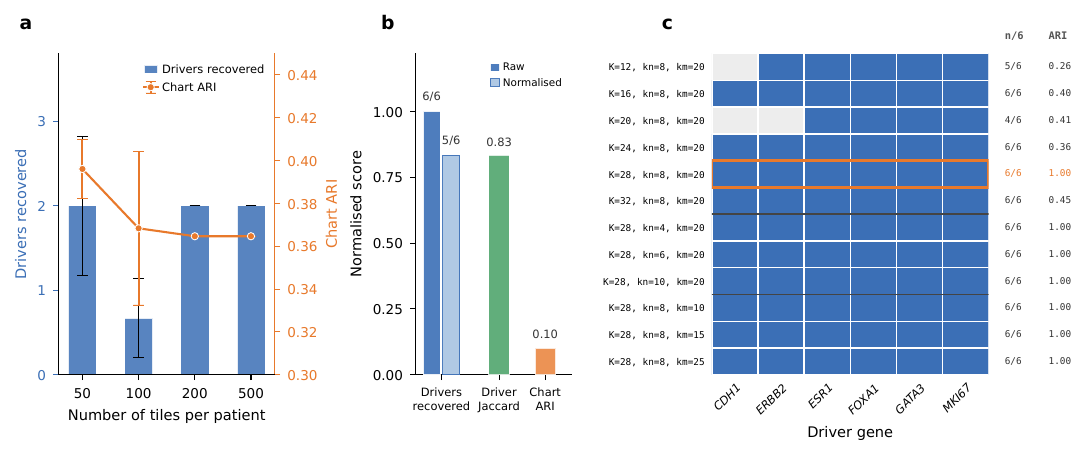}
  \caption{The framework is robust to tile sampling, stain
    normalisation, and hyperparameter variation.}
  \label{fig:convergence}
  \label{fig:sensitivity}
\end{figure}

These results demonstrate that the atlas primarily encodes an
inflammatory--proliferative axis rather than a simple immune-hot
versus immune-cold partition, consistent with the geodesic gene
discovery findings (\cref{fig:deconvolution}).


\subsection{Orthogonal computational pathology grounding}
\label{sec:opentme}

To assess whether atlas coordinates correspond to quantifiable
tissue-composition differences measured by an independent
computational pathology pipeline, we correlated atlas principal
components with slide-level cell segmentation and microenvironment
features from the OpenTME dataset \citep{opentme2024}, which provides
thousands of features per slide derived from automated cell
detection, classification, and spatial neighbourhood analysis of
H\&E whole-slide images.

Of 285 atlas patients, 284 were matched to OpenTME via TCGA case
identifiers.  We computed Spearman rank correlations between each
atlas coordinate (PC1--PC10, PCA-norm) and all available OpenTME
features, applying Benjamini--Hochberg FDR correction across 49{,}973
tests.  A total of 14{,}616 correlations were FDR-significant at
$q < 0.05$ (29.2\% yield).  The strongest individual correlation was
between PC2 and relative carcinoma area ($\rho = 0.652$,
$q = 5.15 \times 10^{-31}$), supporting the interpretation that atlas geometry captures
tumour content variation.

We evaluated four predefined feature-category hypotheses linking atlas coordinates
to specific OpenTME feature categories:
\begin{itemize}
  \item \textbf{H1 (lymphocyte neighbourhood)}: 3{,}186
    FDR-significant correlations out of 9{,}812 tested; top hit:
    lymphocyte neighbourhood co-occurrence ($\rho = -0.496$,
    $q = 4.25 \times 10^{-15}$).
  \item \textbf{H2 (inflammatory cells)}: 3{,}196 FDR-significant
    correlations out of 9{,}812 tested; top hit: macrophage density
    in non-tumour regions ($\rho = 0.471$,
    $q = 2.49 \times 10^{-12}$).
  \item \textbf{H3 (proliferative markers)}: no matching features
    were available in OpenTME, which does not include mitotic-figure
    or Ki-67-based annotations.
  \item \textbf{H4 (cell-type percentages)}: 1{,}861 FDR-significant
    correlations out of 9{,}504 tested; top hit: endothelial cell
    percentage ($\rho = -0.490$, $q = 1.42 \times 10^{-14}$).
\end{itemize}

These results provide orthogonal computational pathology evidence
that atlas states capture compositional differences in the
tumour microenvironment, complementing the RNA-seq-based
deconvolution (\cref{sec:deconvolution}).

\paragraph{Resolution confound control.}
\texttt{IMAGE\_RESOLUTION} correlates strongly with atlas PC1 and PC2
($|\rho| \approx 0.58$), indicating that scanner resolution is a
technical confound.  We repeated all correlations as partial Spearman
tests controlling for \texttt{IMAGE\_RESOLUTION} (rank
residualisation), excluding \texttt{IMAGE\_RESOLUTION} itself from the
test universe (49{,}962 tests; 11 fewer than the original 49{,}973).
Of the 14{,}610 originally significant associations present in this
reduced test set, 6{,}845 (46.9\%) survived after adjustment, while
7{,}765 were lost and 684 new associations emerged (total partial-BH
significant: 7{,}529).  Mean effect size attenuated by 32\%.
Applying hierarchical BH correction (family-level FDR across 25
feature-name families, then within-family correction) yielded
7{,}639 adjusted hits across 24 significant families---higher than
the global count because hierarchical correction is less conservative
when signal concentrates within families.  Per-hypothesis survival
rates were: H1~lymphocyte 49.3\%, H2~inflammatory 54.5\%,
H4~cell-percentage 67.1\%.  After adjustment, atlas PC7 (which has
near-zero resolution correlation, $|\rho| = 0.05$) emerged as the
dominant axis, with macrophage--lymphocyte spatial interactions as the
strongest adjusted signal ($\rho = 0.40$,
$q = 8.4 \times 10^{-9}$).  These results demonstrate that
approximately half the original associations were confounded by
scanner resolution, while the surviving ${\sim}7{,}000$ associations are
resolution-adjusted and consistent with biological signal.  However,
additional confounds (contributing institution, pixel spacing, tissue
area, tumour purity, and staining batch) were not available in the
matched dataset and remain uncontrolled; the adjusted results should
therefore be interpreted as resolution-adjusted rather than fully
scanner-independent, and residual site-level or staining confounding
remains possible.


\subsection{Internal split validation}
\label{sec:internal_split}

To assess the stability of atlas-based gene discovery, we performed
internal split validation using a stratified 70/30 discovery/held-out
split, repeated across three random seeds (CONCH backbone).
Supplementary Table~S5 reports a complementary five-fold 80/20
cross-validation on DINOv2 ViT-B/14; results differ materially
(mean 0.2 drivers per fold vs.\ 3--6 per split here), reflecting
both the different backbone and the smaller per-fold sample size.

\paragraph{Discovery-set atlas.}
Atlases constructed on the 70\% discovery set ($n = 199$) recovered
3--6 FDR-significant drivers per split (split~1: 5/6, split~2: 6/6,
split~3: 3/6), with three drivers (ESR1, FOXA1, GATA3) recovered in
all three splits.

\paragraph{Held-out projection.}
Held-out patients ($n = 86$) were projected onto the discovery-set atlas
by assigning each patient to the nearest chart centroid.
Chart-differential-expression patterns replicated robustly:
$69.0 \pm 4.5$\% of discovery-set chart-DE genes were significant in
the held-out set across three splits (range: 64.9--75.3\%).  Among the
top 50 genes ranked by discovery-set significance, $75.3 \pm 13.9$\%
showed the same direction of effect in the held-out set.  Thirteen
genes replicated at FDR significance in all three splits.  The
Immune\_response pathway replicated in 2/3 splits.

\paragraph{Split stability.}
Ordering-correlation replication was modest ($6.2 \pm 0.1$\%),
consistent with the smaller held-out sample size reducing power for
per-gene ordering correlations.  However, the high directional
concordance confirms that the atlas identifies a consistent set of
morphology-associated genes despite sampling variability
(Supplementary Table~S1).

Chart differential-expression patterns showed partial held-out
replication, while ordering-based gene associations were
substantially less stable (6.2\% replication), indicating that the
atlas captures reproducible chart-level biology but ordering-specific
associations require larger cohorts for robust validation.


\subsection{External molecular replication}
\label{sec:external}

We evaluated the generalisability of atlas-derived gene programmes in
two independent cohorts.  We emphasise that these validations are
\emph{molecular replications} (testing whether the same genes show
expression variation along atlas coordinates), not full
image-genomic validations (which would require paired WSIs and
RNA-seq in the external cohorts).

\paragraph{CPTAC-BRCA molecular replication.}
Using the CPTAC-BRCA cohort ($n = 133$, RNA-seq only), we projected
external expression profiles onto the TCGA-trained atlas by computing
the correlation of each CPTAC patient's expression vector with each
chart's gene-expression centroid.  Of the 3{,}300 genes identified as
atlas-associated in TCGA, 2{,}276 were testable in CPTAC, of which
48.6\% showed concordant expression patterns (same direction of
association with projected atlas coordinates after PC1 axis alignment,
$p < 0.05$).  Among the four drivers present in both cohorts, all
replicated with concordant direction at nominal significance (ERBB2,
$p = 0.011$; ESR1, FOXA1, GATA3, all $p < 10^{-4}$).

\paragraph{TCGA-LUAD cross-cancer transfer.}
To test whether geometric gene discovery generalises beyond breast
cancer, we applied the full pipeline to TCGA-LUAD ($n = 518$) using
the DINOv2 ViT-B/14 backbone with LUAD-specific WSIs and expression
data.  The LUAD atlas recovered 3 of 6 tier-1 LUAD drivers (from a
candidate list including EGFR, KRAS, TP53, STK11, KEAP1, NF1) at FDR
significance, with 4\,984 genes showing chart-level differential
expression (\cref{tab:luad_cross_cancer}).  This cross-cancer
result supports the portability of the geometric decoding framework
beyond breast cancer, though establishing generality requires
validation across additional tissue types and cohorts.

\paragraph{Honest framing.}
We note that neither CPTAC nor LUAD validation constitutes a
complete external replication of the image-genomic atlas, which
would require independently collected, paired WSI and expression
data processed through the full pipeline.  The CPTAC analysis
validates expression-pattern consistency; the LUAD analysis validates
methodological generalisability.  Full image-genomic external
validation remains an important goal for future work.


\subsection{Ablation: geometric contribution to gene discovery}
\label{sec:geometry_ablation}

To isolate the contribution of Riemannian geometry to gene discovery,
we performed a controlled ablation using the DINOv2 ViT-B/14 backbone
with a simplified atlas configuration (16 charts, $k = 6$ neighbours).
We compared four conditions: (i)~full Riemannian pipeline (metric
tensors, geodesic distances); (ii)~Euclidean-only pipeline (L2
distances, identical graph construction and FDR correction);
(iii)~PCA-reduced embeddings (10 components, 85.2\% variance
explained); and (iv)~random baseline (isotropic 10-dimensional
vectors).

The Riemannian and Euclidean pipelines produced identical gene
discovery: both recovered 5/6 canonical drivers (ERBB2, ESR1, FOXA1,
GATA3, MKI67) and 2{,}595 FDR-significant genes across all discovery
modes (\cref{tab:geometry_ablation}).  PCA-10 embeddings also recovered
the same 5/6 drivers (2{,}602 FDR genes), demonstrating that the
low-rank structure of foundation-model embeddings carries the
biological signal.  The random baseline recovered only 1/6 drivers
(MKI67, 659 FDR genes), confirming that real embedding
structure---not statistical artefact---drives gene discovery.
Notably, DINOv2 ViT-S/14 (384-d) recovered 6/6 drivers under the
same ablation configuration (2{,}611 FDR genes), suggesting that
compact architectures may suffice for this cohort size.

These results establish that the Riemannian framework provides
geometric interpretability---curvature hotspot identification,
geodesic path visualization, cocycle consistency checks, and
principled multi-resolution construction---rather than additional
discovery power over Euclidean distances applied to the same
embeddings.  The signal resides in the foundation-model
representations themselves; the geometric machinery organises and
exposes that signal through structured, interpretable constructions.


\section{Discussion}
\label{sec:discussion}

We set out to determine which part of a geometric image-genomic
pipeline carries patient-level molecular signal, and the answer is
narrower than the framing such pipelines usually receive, this one
included, in its earlier form.

Frozen foundation-model embeddings do carry real signal: held-out
Spearman $\rho$ up to $0.556$, against a matched null that re-runs
the same harness on permuted labels.  That result survives every
control we could construct.  But three of the qualifications matter
as much as the headline.  The basal programme is fully explained by
tissue composition, so morphology adds nothing there.  Fifty-four
countable cell features come within $0.043$--$0.085$ of a
1{,}536-dimensional foundation model on every programme.  And the
Riemannian geometry (the element named in the previous title) is a
precise null whose cause is a line of neighbour-selection code rather
than a property of the data.

We think the most useful contribution here is the control suite
itself.  The field's default evidence pattern---cohort-wide gene
ranking, driver counting, and nulls that shuffle structure rather
than compete with it---can report success for a pipeline that a
held-out patient would not validate, and can attribute that success
to a component that provably never influenced the computation.  Each
of those failure modes is present in the earlier version of this
work, and each is detectable with the controls reported here.

\paragraph{Relation to pathology foundation models.}
Recent large-scale models such as Virchow \citep{vorontsov2024virchow},
CHIEF \citep{wang2024chief}, UNI \citep{chen2024uni} and CONCH
\citep{lu2024conch} demonstrate that self-supervised pretraining
on millions of WSIs produces embeddings that transfer well to
diverse downstream tasks.  Our work complements these models by
showing that their representations carry biological signal, while the
geometric apparatus evaluated here adds no measurable patient-level
predictive value.  The multi-resolution analysis further suggests that
biological
information is distributed across spatial scales within these
representations, consistent with hierarchical architectures like
HIPT \citep{chen2022hipt}.

\paragraph{Multi-resolution insights.}
The observation that cellular, mesoscale, and tissue-level atlases
recover partially non-overlapping gene sets has implications for
computational pathology pipeline design.  Current MIL frameworks
typically operate at a single magnification; our results suggest
that multi-scale feature aggregation may improve biological
interpretability.  The fused atlas returned a partly distinct gene
set, but its predictive advantage was not established by the
patient-level benchmark.

\paragraph{Spatially-aware aggregation.}
The comparison of five aggregation strategies showed that
spatially-aware methods (graph attention, top-$k$) returned different
gene sets from mean pooling, but because driver recovery is weakly
discriminative on this cohort (\cref{sec:gene_discovery}), the
differences are descriptive rather than evidence of predictive
improvement.

\paragraph{Why shuffled nulls were not enough.}
The five null models retained from the original analysis (shuffled
linkage, random embeddings, permuted charts, rewired graphs, random
pathways) all test the same proposition: that the pipeline
outperforms \emph{destroyed} structure.  We now regard that as a weak
form of evidence.  Shuffled nulls are nearly always exceeded, because
almost any procedure applied to real data beats the same procedure
applied to noise.  What they cannot answer is the question a reader
actually has: does this pipeline beat a \emph{competing explanation}
that is also plausible and also cheap?

The distinction is not hypothetical here.  Four of the five shuffling
nulls were rejected at $p < 0.05$ (the rewired-graph null was not),
and on that basis the earlier version
of this work concluded that atlas-based discovery was not an
artefact.  Yet the competing-model analysis of
\cref{sec:competing_models} shows that for the basal programme,
tissue composition alone matches the embedding ($p = 0.77$), and the
patient-level ablation of \cref{sec:geometry_patient} shows that the
geometric component contributes nothing at all.  Both facts are
invisible to a shuffling null, and both change what can honestly be
claimed.

The one control in the original analysis of the competing-explanation
type---partialling out scanner resolution from the OpenTME
correlations---is also the one that produced a substantive negative:
53.2\% of significant correlations did not survive it.  That is the
pattern we would emphasise to others building similar pipelines.  A
control is informative in proportion to how plausible the alternative
it tests actually is.

We retain the shuffling nulls as evidence that the harness is not
self-fulfilling, and add a matched permutation null at $10{,}000$
permutations that re-runs the identical fold structure and nested
penalty selection on permuted patient labels, so that any optimism in
the evaluation is present in the null as well as in the observed
statistic.

\paragraph{Embedding dimensionality.}
Quantitative analysis of embedding geometry reveals that frozen
foundation-model backbones occupy a higher-rank subspace (mean
effective rank 128.0) than learned projections (60.6), despite
similar intrinsic dimensionality (12.5 vs.\ 10.3).  Frozen embeddings
from UNI and Phikon are notably anisotropic (isotropy 0.33--0.34),
distributing variance across many dimensions, whereas learned
projections (HES random: effective rank 8.4; HES trained: 18.2) concentrate
variance into few components: the top 5 PCs explain 97\% and 86\%
of variance, respectively.  This representational compression removes
exactly the local geometric variation that the atlas exploits for gene
discovery.  The trained HES projection (5/6 FDR drivers) outperforms
the random HES (1/6), confirming that task-relevant dimensionality
reduction preserves biologically informative structure while arbitrary
compression destroys it.  This observation has practical implications:
atlas construction benefits from high-rank frozen embeddings, and
practitioners should avoid aggressive dimensionality reduction before
geometric analysis.

\paragraph{Why the Riemannian metric is inert.}
An earlier version of this work reported that Riemannian and
Euclidean distance computation on identical embeddings recover the
same driver genes and FDR gene counts, and attributed this to the
signal residing in the foundation-model representations themselves.
That explanation was wrong, and the patient-level ablation
(\cref{sec:geometry_patient}) identifies the actual cause.

The geodesic-graph construction selects neighbours by Euclidean
nearest-neighbour search and only then reweights the edges it has
already selected using the local metric tensor.  The graph topology is therefore Euclidean by
construction: the metric never influences which patients are
neighbours, and so never influences chart membership or anything
downstream of it.  The observed equivalence is a property of the
implementation, not evidence about the data or about the geometry of
foundation-model embedding spaces.  This distinction matters, because
the original phrasing invited the reading that flat and curved
descriptions of these embeddings are empirically indistinguishable---a
claim the ablation never tested.

Two further results follow.  First, the null is \emph{precise} rather
than underpowered: the Riemannian-versus-Euclidean difference is
$+0.0010$ with a 95\% interval of $[-0.0007, +0.0029]$, bounded well
below any effect size that would matter.  Second, when the metric is
applied consistently---selecting neighbours \emph{by} Riemannian
distance rather than reweighting Euclidean-selected edges---the
geometry becomes measurably \emph{worse} ($-0.0117$, 95\% CI
$[-0.0229, -0.0004]$, favoured in only 15 of 44 comparisons).  We
therefore do not claim that Riemannian geometry is irrelevant to
representation learning in general; we claim that this construction
does not implement it, and that the natural consistent implementation
does not help on this task at this cohort size.

We retain the curvature machinery for interpretation only, with the
same caveat as before: the estimator (Christoffel-symbol approximation
with diagonal metric and quadratic Gamma products only) is a
first-order proxy.  Synthetic-manifold validation shows correct sign
discrimination but limited quantitative accuracy (Supplementary Note),
so curvature values are ordinal indicators of metric-tensor variation,
not precise Riemannian invariants.

\paragraph{Cell-type composition.}
The deconvolution analysis revealed that atlas coordinates primarily
capture an inflammatory--proliferative axis ($\rho = 0.379$ for
inflammatory score with PC1; $\rho = -0.231$ for proliferation)
rather than a simple immune-hot versus immune-cold partition.  This
is consistent with the growing recognition that tumour
microenvironment heterogeneity (particularly inflammatory signalling) is
a major axis of morphological variation visible in H\&E
\citep{saltz2018spatial}.  The enrichment of inflammation-associated
charts at the extremes of the morphological ordering suggests that
geometric transition zones may correspond to tumour-immune
boundaries, though this remains a hypothesis requiring spatial
validation.

\paragraph{Computational pathology grounding.}
The OpenTME analysis provides a methodologically orthogonal
re-measurement of the same slides,
distinct from RNA-seq deconvolution, that atlas coordinates track
tissue-composition variation measurable by automated cell
segmentation.  However, scanner resolution
(\texttt{IMAGE\_RESOLUTION}) correlates strongly with atlas PC1/PC2
($|\rho| \approx 0.58$), and controlling for this confound via
partial Spearman correlations reduces the FDR-significant hit count
from 14{,}616 to 6{,}845 (46.8\% survival).  Mean effect sizes
attenuate by 32\%, confirming that roughly half the original signal
was resolution-driven.  Critically, 24 of 25 OpenTME feature families
remain significant after hierarchical FDR correction, and atlas PC7
(which has near-zero resolution correlation) emerges as the dominant
confound-free biological axis, driven by macrophage--lymphocyte
spatial interactions.  Cell-percentage features show the highest
survival rate (67.1\%), while lymphocyte neighbourhood features are
most affected (49.3\% survival).  These resolution-controlled results
indicate that atlas geometry encodes genuine, resolution-adjusted
tissue-composition gradients, while also demonstrating that
approximately half of the raw correlations in multi-site TCGA analyses
should be treated with caution.  Future work should include
resolution-stratified analyses, contributing-site controls, and,
where cell-coordinate data become available, spatially resolved
single-cell neighbourhood analyses.

\paragraph{Clinical caveats.}
We deliberately frame survival results as clinical associations
rather than prognostic claims.  With only 33 OS events, the TCGA-BRCA
cohort is underpowered for survival modelling, and the atlas PC1 does
not add significant information beyond PAM50 subtype.  The survival
association observed for CONCH (C-index = 0.734, log-rank $p =
0.009$) is promising but must be validated in larger,
event-enriched cohorts before any clinical utility can be claimed.

\paragraph{Limitations.}
Several limitations merit discussion.  First, the cohort size ($n =
285$) is modest by genomic standards, and it bounds what the negative
results can mean: the geometry ablation is a precise null \emph{at
this cohort size}, and we cannot exclude that a metric-aware
construction would help with an order of magnitude more patients.
Second, no external image-genomic validation exists.  CPTAC ($n =
133$) has no paired imaging and therefore validates only the
gene-list half of any claim; the segmentation cohort introduced here
has no embeddings, and 284 of its patients are the discovery cohort
itself.  We state plainly that this work contains no external
validation of the atlas, and that none is currently possible with the
data available to us.  Third, mean-pooled tile embeddings may lose
intra-tumour heterogeneity; every result here concerns
patient-averaged morphology.  Fourth, the atlas is constructed
independently per backbone, complicating cross-backbone comparison.
Fifth, cell-type deconvolution relies on expression signatures rather
than single-cell ground truth.  Sixth, multi-resolution analysis
requires raw WSIs.  Seventh, the four gene programmes were frozen in
advance but chosen by us; a different choice could shift which
programmes appear composition-driven, and only basal was tested
against composition with a strong prior that it might fail.

\paragraph{What would change our conclusions.}
We name these in advance rather than after the fact.  A
metric-aware construction that selects neighbours \emph{by} the
learned metric and beats Euclidean on this same held-out benchmark
would overturn \cref{sec:geometry_patient}; our own consistent
implementation does not, but it is one implementation.  A cohort with
paired images and expression for patients disjoint from TCGA-BRCA
would let the benchmark be repeated externally and could move any of
these numbers.  Composition features richer than seven compartment
fractions might close the remaining gap on ER/luminal, proliferation
and immune, which would strengthen rather than weaken the
composition interpretation.  And a demonstration that driver-count
recovery predicts something a held-out patient cares about would
restore the metric we have demoted.

\paragraph{Future directions.}
Several extensions are natural: (i)~repeating this benchmark on a
cohort with no slide overlap with TCGA-BRCA; (ii)~testing whether
metric learning \emph{selected into} the graph construction, rather
than applied after it, recovers the advantage the current
implementation cannot; (iii)~integration with spatial transcriptomics
to ground atlas coordinates in spatially-resolved expression;
(iv)~extension to pan-cancer cohorts; (v)~incorporation of mutation
and copy-number data alongside expression.


\section{Methods}
\label{sec:methods}

\subsection{Cohort and data sources}
\label{sec:methods_cohort}

\paragraph{TCGA-BRCA discovery cohort.}
We analysed 285 patients from The Cancer Genome Atlas Breast Invasive
Carcinoma (TCGA-BRCA) project with matched diagnostic
formalin-fixed, paraffin-embedded (FFPE) whole-slide H\&E images and
RNA-seq gene expression profiles (RSEM-normalised, upper-quartile
normalised, $\log_2(x + 1)$ transformed).  Clinical annotations
included age at diagnosis, overall survival (OS) status and time,
disease-free survival (DFS) status and time, oestrogen-receptor (ER)
status, and PAM50 intrinsic subtype.

\paragraph{CPTAC-BRCA replication cohort.}
Molecular replication was performed using 133 patients from the
Clinical Proteomic Tumour Analysis Consortium (CPTAC) breast cancer
cohort (RNA-seq expression only, no paired WSIs).

\paragraph{TCGA-LUAD cross-cancer cohort.}
Cross-cancer generalisability was assessed using 518 TCGA-LUAD
patients with matched WSIs and RNA-seq profiles.

\paragraph{Segmentation-derived cohort and its independence boundary.}
Under the Aignostics Research Access Programme we additionally
obtained OpenTME spatial segmentation outputs for 1{,}125 TCGA-BRCA
slides from 1{,}054 patients (cell polygons, compartment polygons,
raster maps and per-cell readouts; 1.23\,TB).  We state the boundary
explicitly because it constrains every claim made from this resource.
Of these patients, 284 are the same patients used for atlas
discovery, and 770 are not (1{,}054 $-$ 284 $=$ 770; six of these
were subsequently excluded for missing clinical annotations, leaving
764 in the independent analysis set, of which 761 have GDC
STAR-Counts RNA-seq).  The programme supplies
\emph{segmentation, not representations}: no embeddings, model
weights, layers or pooled features are provided for any patient.
Consequently no full-atlas external validation is possible, since the
independent patients have no embeddings from which an atlas could
be built.  Analyses using this resource are described as
segmentation-derived composition analyses and are never described as
external validation of the atlas.  Where the resource overlaps the
discovery cohort it shares the same slides, so it is a
re-measurement, not an independent sample.

\subsection{Patient-level held-out benchmark}
\label{sec:methods_patient_benchmark}

All primary results in this revision are produced by a single
harness, applied identically to every model family so that
comparisons are paired.

\paragraph{Unit and splitting.}
The patient is the independent unit throughout.  Predictions are
out-of-fold under \texttt{GroupKFold} with five folds, grouped by
patient, so no patient contributes to both the fit and the evaluation
of any prediction.  All preprocessing---imputation, standardisation,
dimensionality reduction, and penalty selection by internal
cross-validation---is fitted inside the training fold only.  Folds are
constructed once and reused across every model family, so all
contrasts are paired on identical splits.

\paragraph{Targets.}
Four gene programmes were frozen before any model was fitted:
ER/luminal, proliferation, basal, and immune.  Programme scores are
computed by $z$-scoring each member gene across patients and
averaging, the standard signature-score construction.  Programme
membership is listed in
\texttt{03\_BASELINES/frozen\_programmes.json} and was not revised
after inspecting any result.

\paragraph{Metric.}
Spearman $\rho$ between out-of-fold predictions and measured
programme score, computed once over all patients rather than averaged
across folds.

\paragraph{Permutation null.}
The null permutes patient programme scores and re-runs the
\emph{identical} harness---the same folds, the same nested penalty
selection, the same preprocessing---so that any optimism introduced
by the harness is present in the null as well as in the observed
statistic.  Primary nulls use $10{,}000$ permutations.  We note that
an earlier iteration of this analysis used 200 permutations, at which
every cell reported the floor value $p = 0.005$ and no cell could be
distinguished from any other; the $10{,}000$-permutation results
supersede it.  Reported $p$-values are $(1 + \#\{{\rm null} \geq
{\rm observed}\}) / (1 + n_{\rm perm})$, and family-wise correction
across the 44 primary cells is by Holm--Bonferroni.

\paragraph{Competing model families.}
Every programme is predicted from each of: an intercept-only floor;
tissue composition alone (compartment relative areas); technical
covariates alone (scanner resolution, tissue area, artefact, focus
and marker quality terms); interpretable cell-count features; the
mean-pooled backbone embedding; and embedding plus composition
concatenated.  The composition, technical and cell-count features are
taken from the OpenTME root tables and aggregated to the patient by
averaging across that patient's slides.

\subsection{Tile extraction and embedding}
\label{sec:methods_embedding}

\paragraph{Tile extraction.}
For each WSI, non-overlapping tiles of $224 \times 224$ pixels were
extracted at the second pyramid level (${\sim}20{\times}$ equivalent
magnification on standard $40{\times}$ scanners).  Tissue content was
assessed per tile by computing the mean normalised grayscale intensity
and requiring that at least 70\% of pixels fall below an intensity
threshold of 0.85 (on a 0--1 scale); tiles not meeting this criterion
were discarded.  For multi-resolution analysis, tiles of the same
$224 \times 224$\,px size were extracted at pyramid levels
corresponding to ${\sim}10{\times}$ and ${\sim}5{\times}$
magnification, so that each tile captures a progressively larger
field of view.

\paragraph{Foundation model backbones.}
Tile images were encoded using nine frozen pretrained backbones:
\begin{itemize}
  \item DINOv2 ViT-S/14 (384-dim), ViT-B/14 (768-dim), ViT-L/14
    (1024-dim) \citep{oquab2024dinov2}
  \item Phikon (768-dim), Phikon-v2 (1024-dim) \citep{filiot2023phikon}
  \item UNI (1024-dim) \citep{chen2024uni}
  \item CONCH (512-dim) \citep{lu2024conch}
  \item DINOv2+HES random projection (32-dim), DINOv2+HES trained
    projection (32-dim)
\end{itemize}
All models were used in inference mode without fine-tuning.  Images
were preprocessed according to each model's standard protocol
(centre crop, ImageNet normalisation for DINOv2; model-specific
normalisation for pathology models).

\paragraph{Patient-level aggregation.}
By default, tile embeddings were mean-pooled to produce a single
patient-level vector.  For spatially-aware aggregation experiments,
we additionally computed gated-attention, graph-attention, and
top-$k$ pooled representations (see \cref{sec:methods_aggregation}).

\paragraph{Additional backbones.}
Where publicly available model weights permitted, we additionally
evaluated Virchow \citep{vorontsov2024virchow}, CHIEF
\citep{wang2024chief}, and UNI-v2.

\subsection{Multi-resolution embedding extraction}
\label{sec:methods_multiresolution}

For multi-resolution analysis, embeddings were extracted at three
spatial scales:
\begin{itemize}
  \item \textbf{Cellular}: $224 \times 224$\,px tiles at the pyramid
    level closest to $20{\times}$ magnification, capturing nuclear
    and cell-level morphology.
  \item \textbf{Mesoscale}: $224 \times 224$\,px tiles at the pyramid
    level closest to $10{\times}$ magnification, capturing glandular
    patterns within a broader field of view.
  \item \textbf{Tissue}: $224 \times 224$\,px tiles at the pyramid
    level closest to $5{\times}$ magnification, capturing
    architectural patterns across large tissue regions.
\end{itemize}
Each scale produced an independent set of tile embeddings, which were
mean-pooled per patient.  The fused representation was obtained by
concatenating the $\ell_2$-normalised patient-level vectors from all
three scales and applying PCA to reduce the concatenated vector to
the dimensionality of the cellular-scale embedding.

\subsection{Riemannian atlas construction}
\label{sec:methods_atlas}

Atlas construction proceeds in five stages:

\paragraph{Chart construction.}
Patient-level embeddings $\{\mathbf{x}_i\}_{i=1}^N$, $\mathbf{x}_i
\in \mathbb{R}^d$, were partitioned into $K$ overlapping coordinate
charts using spherical $k$-means clustering.  Each chart $\mathcal{C}_k$
contains all patients within a radius $r_k$ of the chart centroid
$\boldsymbol{\mu}_k$, where $r_k$ is set to include all assigned
members plus a 20\% overlap margin.  The chart ordering value
$\bar{o}_k$ is the PCA-norm of each patient's embedding: the $L_2$
norm of the first three principal components of the patient-level
embedding matrix, normalised to $[0, 1]$.  This ordering is computed
entirely from image embeddings without any clinical or molecular
input.  The per-chart ordering $\bar{o}_k$ is the mean of the
member-level ordering values within the chart.

\paragraph{Graph construction.}
A $k$-nearest-neighbour graph $\mathcal{G} = (\mathcal{V},
\mathcal{E})$ was constructed over chart centroids in embedding space,
with $k = 8$ neighbours.  Edge weights were Euclidean distances between
centroids.

\paragraph{Metric tensor estimation.}
Within each chart, a local Riemannian metric tensor $G_k \in
\mathbb{R}^{d' \times d'}$ was estimated from the covariance of
local embedding coordinates (patient positions relative to chart
centroid), where $d'$ is the intrinsic dimensionality estimated by
PCA.  Geodesic distances between charts were computed as
$d_\text{geo}(i, j) = \|\boldsymbol{\mu}_i - \boldsymbol{\mu}_j
\|_{G_{ij}}$, where $G_{ij} = (G_i + G_j) / 2$ is the arithmetic mean of the endpoint
metric tensors (a first-order approximation to the Fr\'echet mean on
the SPD manifold).

\paragraph{Curvature estimation.}
Scalar curvature at each chart was estimated via Christoffel-symbol
approximation in local PCA coordinates.  For each chart $k$, a
kernel-weighted local covariance matrix was computed at positions
displaced along each coordinate axis; central finite differences of
these covariance matrices yielded the three metric-gradient terms
$\partial g_{ab}/\partial x^c$, from which Christoffel symbols and
the Riemann curvature tensor were assembled.  Curvature magnitudes
are expressed in PCA-coordinate units and scale with embedding
dimension; within a given backbone, charts with positive curvature
($\bar{\kappa} > 0$) identify regions where the metric tensor
changes most rapidly, corresponding to geometric transition zones.

\paragraph{Geodesic path extraction.}
Energy-minimising paths between all pairs of ``terminal'' charts
(charts in the lowest and highest ordering-value quintiles) were
extracted via Dijkstra's algorithm on the metric-weighted graph.
Paths were classified as monotonic if the Spearman correlation between
chart position along the path and the image-derived ordering (PCA-norm)
exceeded $|\rho| > 0.7$ with $p < 0.05$.

\subsection{Manifold-guided gene discovery}
\label{sec:methods_genes}

Three complementary discovery modes were applied:

\paragraph{Transition genes.}
For each edge $(i, j) \in \mathcal{E}$, a two-sided Mann--Whitney
$U$ test was applied to each gene, comparing expression in chart $i$
vs.\ chart $j$.  Resulting $p$-values were corrected across all genes
and all edges using the Benjamini--Hochberg (BH) procedure at $q <
0.05$.

\paragraph{Monotonic genes.}
For each geodesic path, Spearman rank correlation between gene
expression and chart position along the path was computed.  $p$-values
were BH-corrected across all genes per path.

\paragraph{Correlation-ranked genes.}
Spearman correlation between gene expression and the first principal
component (PC1) of the atlas embedding was computed.  $p$-values were
BH-corrected across all genes.

\paragraph{Driver-gene evaluation.}
Recovery of six canonical breast-cancer driver genes (ESR1, FOXA1,
GATA3, CDH1, MKI67, ERBB2) is retained as a descriptive historical
metric; all primary comparative conclusions use held-out patient-level
programme prediction (\cref{sec:patient_benchmark}).  A driver was
counted as ``recovered'' if it appeared among the FDR-significant
genes in any discovery mode.

\paragraph{Pathway enrichment (Path-GSEA).}
For each geodesic path, MSigDB Hallmark gene-set scores
\citep{liberzon2015molecular} were computed per patient (mean
expression of set members) and tested for monotonic trends along the
path using Spearman correlation with BH correction.

\subsection{Spatially-aware tile aggregation}
\label{sec:methods_aggregation}

Five aggregation strategies were compared:

\paragraph{Mean pooling.}
$\bar{\mathbf{x}} = \frac{1}{T} \sum_{t=1}^T \mathbf{x}_t$.

\paragraph{Gated attention pooling.}
Attention weights $a_t = \text{softmax}(\mathbf{w}^\top
\tanh(\mathbf{V}\mathbf{x}_t) \odot
\text{sigm}(\mathbf{U}\mathbf{x}_t))$ were computed, and the
patient representation was $\bar{\mathbf{x}} = \sum_t a_t
\mathbf{x}_t$ \citep{ilse2018attention}.  Parameters
$\mathbf{w}, \mathbf{V}, \mathbf{U}$ were set by norm-based
initialisation (no training).

\paragraph{Graph attention pooling.}
A $k$-NN graph ($k = 8$) was constructed over tile spatial coordinates
(centroid positions in the WSI).  Two hops of message passing were
applied, where each tile aggregated features from its spatial
neighbours weighted by cosine similarity in embedding space.
Attention-weighted readout produced the patient-level vector.

\paragraph{Top-$k$ pooling.}
The 50 tiles with highest embedding-norm variance were selected, and
their embeddings were mean-pooled.

\paragraph{Coordinate-aware pooling.}
Tile embeddings were weighted by a spatial kernel (Gaussian, $\sigma
=$ slide width $/4$) centred on the slide centroid, downweighting
peripheral tiles.

\subsection{Visual grounding}
\label{sec:methods_grounding}

\paragraph{Prototype retrieval.}
For each atlas chart, the five patients closest to the chart centroid
in embedding space were identified.  For each patient, tile-level
embeddings were loaded and the tile whose embedding was nearest to
the patient mean was selected.  The corresponding tile image was
extracted from the TCGA-BRCA whole-slide image (SVS format) via
OpenSlide, using the same pyramid level and tissue-filtered grid as
the original tile extraction pipeline.  Tiles failing a tissue
fraction threshold ($> 70\%$ of pixels below 0.85 normalised
intensity) were excluded.

\paragraph{Attention rollout.}
For the DINOv2 ViT-B/14 backbone, attention-rollout maps were
computed by multiplying attention weight matrices across all
transformer layers.  The resulting spatial attention map was overlaid
on each prototype tile as a heatmap, highlighting sub-tile regions
that contributed most to the patient embedding.

Visual interpretations are qualitative; pathologist-reviewed labels
were not available for the present analysis.

\subsection{Stain normalisation}
\label{sec:methods_stain}

Three stain-normalisation conditions were compared:

\begin{enumerate}
  \item \textbf{None}: raw tile images.
  \item \textbf{Reinhard} \citep{reinhard2001color}: global colour
    transfer matching the mean and standard deviation of each Lab
    channel to a reference slide.
  \item \textbf{Macenko} \citep{macenko2009method}: stain
    deconvolution via singular-value decomposition, followed by
    restaining with reference stain vectors.
\end{enumerate}

For each condition, tile embeddings were re-extracted, patient-level
representations recomputed, and the atlas rebuilt.  Driver-gene
recovery and atlas topology (Adjusted Rand Index of chart
assignments) were compared across conditions.

\subsection{Null models}
\label{sec:methods_nulls}

Five null models were constructed, each with 1{,}000 permutations:

\begin{enumerate}
  \item \textbf{Shuffled linkage}: expression profiles were randomly
    permuted across patients while keeping image embeddings fixed.
  \item \textbf{Random embeddings}: patient embeddings were replaced
    by isotropic Gaussian vectors with matched mean, variance, and
    dimensionality.
  \item \textbf{Permuted charts}: patient-to-chart assignments were
    randomly permuted while preserving the graph structure.
  \item \textbf{Rewired graph}: chart adjacency was randomly rewired
    using the Maslov--Sneppen algorithm, preserving degree distribution.
  \item \textbf{Random pathways}: geodesic paths were extracted from
    Erd\H{o}s--R\'enyi random graphs with matched edge density.
\end{enumerate}

For each null permutation, the full gene-discovery pipeline was
rerun and the number of recovered FDR-significant driver genes was
recorded.  The empirical $p$-value was computed with the standard
plus-one correction $p = (b + 1) / (B + 1)$, where $b$ is the
number of null permutations achieving driver recovery $\geq$ the
observed value and $B = 1{,}000$ is the total number of permutations.

\subsection{Bootstrap survival analysis}
\label{sec:methods_survival}

Survival analysis was performed using multivariable Cox
proportional-hazards models with three covariates: atlas PC1
(continuous, 0--1 range), age at diagnosis (standardised), and ER
status (binary).  The concordance index (C-index) was estimated
from 1\,000 bootstrap resamples with bias-corrected and accelerated
(BCa) 95\% confidence intervals.  Kaplan--Meier curves were
stratified by PC1 tertiles, and log-rank tests were computed for
the comparison of lowest vs.\ highest tertiles.

Model comparison was performed by nested likelihood-ratio tests:
Model~0 (age + ER), Model~1 (age + ER + PAM50), Model~2 (age + ER +
atlas PC1), Model~3 (age + ER + PAM50 + atlas PC1).  The
contribution of atlas PC1 beyond PAM50 was assessed by the
likelihood-ratio test comparing Model~3 to Model~1.

\subsection{Cell-type deconvolution}
\label{sec:methods_deconvolution}

Cell-type composition was estimated using single-sample gene-set
enrichment analysis (ssGSEA) with curated gene-expression signatures
for the following cell types and programmes:

\begin{itemize}
  \item \textbf{Immune}: CD8+ T cells, macrophages, B cells, NK cells
    (LM22 signatures \citep{newman2015robust}).
  \item \textbf{Stromal}: cancer-associated fibroblasts, endothelial
    cells (ESTIMATE signatures \citep{yoshihara2013inferring}).
  \item \textbf{Proliferation}: MKI67-associated programme.
\end{itemize}

For each signature, the enrichment score was computed per patient
and correlated with atlas PC1 and PC2 (Spearman rank correlation).

\subsection{Internal split validation}
\label{sec:methods_split}

The 285-patient cohort was split into a 70\% discovery set ($n
\approx 200$) and a 30\% held-out set ($n \approx 85$), stratified
by ER status.  This split was repeated with three random seeds.

For each split, the atlas was constructed on the discovery set and
gene discovery was performed.  Held-out patients were projected onto
the discovery-set atlas by assigning each to the nearest chart
centroid (Euclidean distance in embedding space).  Gene-expression
patterns along atlas paths were then evaluated in the held-out set
to assess concordance with discovery-set findings.

Split stability was quantified by the Adjusted Rand Index (ARI) of
chart assignments across splits and by the Spearman correlation of
gene-level $p$-value rankings between discovery sets.

\subsection{Parameter sensitivity analysis}
\label{sec:methods_sensitivity}

Parameter sensitivity was evaluated using the DINOv2 ViT-B/14
backbone with baseline $K = 28$, $k_{\text{graph}} = 8$,
$k_{\text{metric}} = 20$.  Axis sweeps varied $K \in \{12, 16, 20,
24, 28, 32\}$, $k_{\text{graph}} \in \{4, 6, 8, 10\}$, and
$k_{\text{metric}} \in \{10, 15, 20, 25\}$ (12 configurations
total).  For each configuration, the atlas was constructed on the
full cohort and driver-gene recovery and chart Adjusted Rand Index
(ARI) relative to baseline were recorded.

\subsection{Embedding dimensionality analysis}
\label{sec:methods_dimensionality}

To understand why learned low-dimensional projections (DINOv2+HES,
32-dim) perform comparably to full-dimensional backbones, we computed:

\begin{itemize}
  \item \textbf{Intrinsic dimensionality}: estimated by the maximum
    likelihood estimator \citep{levina2004maximum} on the full
    embedding set.
  \item \textbf{Effective rank}: the Shannon entropy of the normalised
    singular values of the embedding matrix \citep{roy2007effective}.
  \item \textbf{Isotropy}: the ratio of the minimum to maximum
    singular values, measuring how uniformly the embedding space is
    utilised.
  \item \textbf{Representation collapse}: the fraction of variance
    explained by the top-$d'$ principal components, where $d' =$ the
    intrinsic dimensionality.
\end{itemize}

These metrics were compared across eight backbones (four frozen:
DINOv2 ViT-B/14, CONCH, Phikon, UNI; four learned: JEPA, attention
MIL, HES random, HES trained) to characterise the geometric structure
that underlies atlas-based gene discovery.

\subsection{Software and reproducibility}
\label{sec:methods_software}

All analyses were implemented in Python 3.12 using NumPy, SciPy,
scikit-learn, and matplotlib.  Deep learning models were run via
PyTorch with the \texttt{timm} and \texttt{transformers} libraries.
WSI processing used OpenSlide.  Persistent homology used GUDHI.
Code, pre-computed embeddings, and a computational reproducibility
capsule are available at \url{https://github.com/ChimdiWalter/Pathology_Atlas}.


\section*{Data availability}
\label{sec:data}

TCGA-BRCA and TCGA-LUAD data are available through the Genomic Data
Commons (\url{https://portal.gdc.cancer.gov}).  CPTAC-BRCA data are
available through the Proteomic Data Commons
(\url{https://pdc.cancer.gov}).  OpenTME slide-level tumour
microenvironment features are available at
\url{https://huggingface.co/datasets/Aignostics/OpenTME}.

Additional OpenTME spatial segmentation outputs (cell and compartment
polygons, raster maps, per-cell readouts; 1{,}125 TCGA-BRCA slides
from 1{,}054 patients) were obtained under the Aignostics Research
Access Programme.  Under the terms of that programme these raw
outputs may not be redistributed by us; access is available directly
from Aignostics.  The derived patient-level composition, technical
and cell-count feature tables used in
\cref{tab:competing_models}, which contain no redistributable raw
segmentation, are included in the deposit below.

Pre-computed patient-level embeddings, out-of-fold predictions for
every model family, the full permutation-null distributions, and the
analysis workspace underlying \cref{tab:patient_benchmark,%
tab:competing_models,tab:geometry_patient} will be deposited on
Zenodo upon publication.


\section*{Code availability}
\label{sec:code}

All source code for atlas construction, gene discovery, statistical
validation, and figure generation is available at
\url{https://github.com/ChimdiWalter/Pathology_Atlas}.


\section*{Acknowledgements}

The results shown here are in part based upon data generated by The
Cancer Genome Atlas managed by the NCI and NHGRI.  Information about
TCGA can be found at \url{https://www.cancer.gov/tcga}.  The authors
thank the CPTAC programme and the National Cancer Institute Clinical
Proteomic Tumour Analysis Consortium.  We thank Aignostics GmbH for
providing access to additional OpenTME spatial segmentation outputs
through their Research Access Programme, and for making the OpenTME
dataset publicly available on HuggingFace.


\section*{Author contributions}

C.W.N.\ conceived the study, developed the methodology, wrote all software, performed the experiments and analyses, and wrote the manuscript.


\section*{Competing interests}

The authors declare no competing interests.


\clearpage

\clearpage

\begin{table}[!ht]
\centering
\caption{Multi-resolution atlas: driver-gene recovery across spatial
  scales and backbones.}
\label{tab:multiresolution}
\footnotesize
\begin{tabular}{llrrl}
\toprule
Backbone & Resolution & FDR genes & Drivers & Drivers recovered \\
\midrule
CONCH & Cellular (${\sim}20{\times}$) & 1{,}846 & 6/6 & all six \\
 & Mesoscale (${\sim}10{\times}$) & 1{,}954 & 6/6 & all six \\
 & Tissue (${\sim}5{\times}$) & 1{,}927 & 6/6 & all six \\
 & Fused & 1{,}918 & 6/6 & all six \\
\midrule
DINOv2 ViT-B/14 & Cellular (${\sim}20{\times}$) & 1{,}892 & 5/6 & ERBB2, ESR1, FOXA1, GATA3, MKI67 \\
 & Mesoscale (${\sim}10{\times}$) & 1{,}981 & 6/6 & all six \\
 & Tissue (${\sim}5{\times}$) & 1{,}824 & 6/6 & all six \\
 & Fused & 2{,}029 & 6/6 & all six \\
\bottomrule
\end{tabular}
\\[2pt]
{\footnotesize FDR genes: union of transition, monotonic, and
correlation-ranked genes (BH $q < 0.05$).  Drivers: out of six
canonical breast-cancer genes (ESR1, FOXA1, GATA3, CDH1, MKI67, ERBB2).}
\end{table}

\begin{table*}[!ht]
\centering
\caption{Backbone comparison with Benjamini--Hochberg FDR correction
  ($q < 0.05$).  Nine backbones spanning general-purpose (DINOv2) and
  histopathology-specialised (Phikon, UNI, CONCH) architectures.}
\label{tab:backbone_comparison_fdr}
\footnotesize
\begin{tabular}{l@{\hskip 6pt}r@{\hskip 6pt}r@{\hskip 6pt}r@{\hskip 6pt}r@{\hskip 6pt}r@{\hskip 6pt}r@{\hskip 6pt}r@{\hskip 6pt}r}
\toprule
Backbone & Dim & $n$ & Trans (FDR) & Mono (FDR) & Corr (FDR) & ChDE (FDR) & Drivers & $\bar{\kappa}$ \\
\midrule
DINOv2 ViT-S/14 & 384 & 285 & 0 & 377 & 1\,505 & 894 & 3/6 & $7.8 \times 10^{3}$ \\
DINOv2 ViT-B/14 & 768 & 285 & 0 & 532 & 1\,622 & 942 & 4/6 & $1.2 \times 10^{4}$ \\
DINOv2 ViT-L/14 & 1\,024 & 285 & 0 & 420 & 1\,311 & 5 & 2/6 & $3.4 \times 10^{4}$ \\
Phikon & 768 & 285 & 0 & 283 & 1\,541 & 10 & 3/6 & $2.8 \times 10^{4}$ \\
Phikon-v2 & 1\,024 & 285 & 94 & 265 & 1\,617 & 0 & 4/6 & $2.1 \times 10^{4}$ \\
UNI & 1\,024 & 285 & 50 & 259 & 1\,455 & 0 & 4/6 & $4.1 \times 10^{5}$ \\
CONCH & 512 & 285 & 50 & 433 & 1\,493 & 146 & 3/6 & $1.3 \times 10^{3}$ \\
DINOv2+HES & 32 & 285 & 0 & 429 & 1\,166 & 377 & 1/6 & $2.8 \times 10^{2}$ \\
DINOv2+HES trained & 32 & 285 & 0 & 503 & 1\,693 & 200 & 5/6 & $7.0 \times 10^{2}$ \\
\bottomrule
\end{tabular}
\\[2pt]
{\footnotesize Trans: transition genes. Mono: monotonic genes. Corr:
correlation-ranked genes. ChDE: chart differential expression genes.
Drivers: out of six canonical breast-cancer driver genes
(ESR1, ERBB2, CDH1, FOXA1, GATA3, MKI67) at BH $q < 0.05$.
$\bar{\kappa}$: mean scalar curvature in PCA coordinates;
magnitude reflects embedding dimension and coordinate scale.
Driver counts in this table use the default single-resolution
analysis ($K{=}28$ charts, mean-pooled embeddings); the
multi-resolution table (\cref{tab:multiresolution}) additionally
fuses cellular, mesoscale, and tissue scales, which increases
driver recovery (e.g., CONCH: 3/6 single-resolution $\to$ 6/6
fused).}
\end{table*}

\begin{table*}[!ht]
\centering
\caption{Tile aggregation method comparison: driver-gene recovery from
  atlas structure across three backbones.  Graph attention uses $k$-NN
  feature graphs with message passing.}
\label{tab:tile_aggregation}
\footnotesize
\begin{tabular}{llrrrl}
\toprule
Backbone & Aggregation & Genes & Drivers & $n$/6 & Drivers recovered \\
\midrule
CONCH & Mean (stored) & 1\,506 & 4 & 4/6 & ESR1, FOXA1, GATA3, MKI67 \\
 & Mean (tiles) & 1\,250 & 5 & 5/6 & CDH1, ESR1, FOXA1, GATA3, MKI67 \\
 & Attention & 2\,037 & 5 & 5/6 & CDH1, ESR1, FOXA1, GATA3, MKI67 \\
 & Graph attn. & 1\,699 & 5 & 5/6 & CDH1, ESR1, FOXA1, GATA3, MKI67 \\
 & Top-$k$ & 1\,836 & 5 & 5/6 & CDH1, ESR1, FOXA1, GATA3, MKI67 \\
\midrule
DINOv2 ViT-B/14 & Mean (stored) & 2\,019 & 5 & 5/6 & CDH1, ESR1, FOXA1, GATA3, MKI67 \\
 & Mean (tiles) & 1\,939 & 4 & 4/6 & ERBB2, ESR1, FOXA1, GATA3 \\
 & Attention & 1\,868 & 4 & 4/6 & CDH1, ESR1, FOXA1, GATA3 \\
 & Graph attn. & 1\,764 & 5 & 5/6 & CDH1, ESR1, FOXA1, GATA3, MKI67 \\
 & Top-$k$ & 1\,665 & 5 & 5/6 & ERBB2, ESR1, FOXA1, GATA3, MKI67 \\
\midrule
Phikon & Mean (stored) & 2\,143 & 4 & 4/6 & ESR1, FOXA1, GATA3, MKI67 \\
 & Mean (tiles) & 1\,022 & 5 & 5/6 & ERBB2, ESR1, FOXA1, GATA3, MKI67 \\
 & Attention & 816 & 2 & 2/6 & ERBB2, ESR1 \\
 & Graph attn. & 978 & 3 & 3/6 & ESR1, FOXA1, GATA3 \\
 & Top-$k$ & 1\,847 & 3 & 3/6 & ESR1, FOXA1, GATA3 \\
\bottomrule
\end{tabular}
\\[2pt]
{\footnotesize Mean (stored): pre-computed mean-pooled patient
embeddings. Mean (tiles): re-pooled from tiles. Attention: norm-based
gated attention. Graph attn.: 8-NN cosine graph, 2-hop message
passing, attention pooling. Top-$k$: average of 50 highest-variance
tiles. All gene tests BH $q < 0.05$.}
\end{table*}

\begin{table}[!ht]
\centering
\caption{Path-level gene set enrichment (Path-GSEA) across backbones.
  Pathway-level testing reduces the multiple-testing burden from
  ${\sim}5\,000$ genes to ${\sim}50$ pathways per path.}
\label{tab:path_gsea}
\footnotesize
\begin{tabular}{lrrrrr}
\toprule
 & Mono. & \multicolumn{2}{c}{Path-GSEA} & \multicolumn{2}{c}{Gene-level FDR} \\
\cmidrule(lr){3-4} \cmidrule(lr){5-6}
Backbone & paths & Sig.\ paths & Pathways & Sig.\ genes & Drivers \\
\midrule
DINOv2 ViT-S/14 & 9 & 1 & 5 & --- & 0/6 \\
DINOv2 ViT-B/14 & 9 & 0 & 0 & --- & 0/6 \\
DINOv2 ViT-L/14 & 9 & 0 & 0 & --- & 0/6 \\
Phikon & 9 & 3 & 9 & --- & 4/6 \\
Phikon-v2 & 9 & 1 & 1 & --- & 1/6 \\
UNI & 5 & 0 & 0 & --- & 0/6 \\
CONCH & 9 & 3 & 21 & --- & 2/6 \\
DINOv2+HES & 9 & 1 & 1 & --- & 0/6 \\
DINOv2+HES trained & 9 & 1 & 3 & --- & 0/6 \\
\bottomrule
\end{tabular}
\\[2pt]
{\footnotesize Path-GSEA: Spearman correlation of pathway scores
(mean expression of gene-set members) vs.\ path ordering, BH $q <
0.05$.  Gene-level: individual Spearman tests with BH correction.}
\end{table}

\begin{table}[!ht]
\centering
\caption{Curvature hotspot characterisation: genes differentially
  expressed in high-curvature (morphological transition) regions and
  their Hallmark pathway enrichment.}
\label{tab:curvature_hotspot}
\footnotesize
\begin{tabular}{lrrrrl}
\toprule
Backbone & Curv.\ genes & DE genes & Fisher paths & PS DE & Key programmes \\
\midrule
CONCH & 1\,221 & 0 & 0 & 0 & --- \\
Phikon & 1\,111 & 862 & 1 & 16 & TNF$\alpha$ signalling \\
Phikon-v2 & 630 & 0 & 0 & 1 & --- \\
UNI & 30 & 0 & 0 & 0 & --- \\
DINOv2 ViT-S/14 & 93 & 305 & 4 & 1 & Apical junction, TNF$\alpha$ \\
DINOv2 ViT-B/14 & 386 & 119 & 2 & 7 & Apical junction, P53 pathway \\
DINOv2 ViT-L/14 & 46 & 523 & 5 & 14 & E2F targets, TNF$\alpha$ \\
\bottomrule
\end{tabular}
\\[2pt]
{\footnotesize Curv.\ genes: Mann--Whitney $p < 0.01$.  DE genes:
density-gradient proxy, BH $q < 0.05$.  Fisher: one-sided,
BH-corrected.  PS DE: pathway score rank-sum, BH $q < 0.05$.}
\end{table}

\begin{table*}[!ht]
\centering
\caption{Bootstrap survival analysis with 95\% CIs across nine
  backbones.  Multivariable Cox PH (PC1 + age + ER status).  Only
  CONCH achieves significance.}
\label{tab:bootstrap_survival}
\footnotesize
\begin{tabular}{l@{\hskip 4pt}r@{\hskip 4pt}r@{\hskip 6pt}l@{\hskip 6pt}r@{\hskip 4pt}r@{\hskip 4pt}r@{\hskip 6pt}l}
\toprule
 & \multicolumn{3}{c}{Overall survival} & & \multicolumn{3}{c}{Disease-free survival} \\
\cmidrule(lr){2-4} \cmidrule(lr){6-8}
Backbone & C-index & 95\% CI & $p_\text{LR}$ & & C-index & 95\% CI & $p_\text{LR}$ \\
\midrule
DINOv2 ViT-S/14 & 0.739 & [0.639, 0.822] & 0.111 & & 0.610 & [0.485, 0.739] & 0.695 \\
DINOv2 ViT-B/14 & 0.740 & [0.634, 0.819] & 0.201 & & 0.612 & [0.476, 0.737] & 0.550 \\
DINOv2 ViT-L/14 & 0.738 & [0.636, 0.816] & 0.261 & & 0.611 & [0.476, 0.737] & 0.183 \\
Phikon & 0.728 & [0.635, 0.810] & 0.148 & & 0.619 & [0.495, 0.741] & 0.957 \\
Phikon-v2 & 0.731 & [0.636, 0.810] & 0.405 & & 0.619 & [0.493, 0.741] & 0.785 \\
UNI & 0.728 & [0.635, 0.810] & 0.105 & & 0.619 & [0.497, 0.741] & 0.723 \\
CONCH & \textbf{0.734} & [0.638, 0.820] & \textbf{0.009} & & 0.623 & [0.501, 0.739] & 0.322 \\
DINOv2+HES & 0.737 & [0.642, 0.814] & 0.879 & & 0.616 & [0.487, 0.734] & 0.842 \\
DINOv2+HES trained & 0.740 & [0.633, 0.821] & 0.167 & & 0.614 & [0.481, 0.738] & 0.334 \\
\bottomrule
\end{tabular}
\\[2pt]
{\footnotesize C-index: median of 1\,000 bootstrap resamples.
$p_\text{LR}$: log-rank test on atlas PC1 tertiles.  Bold = $p < 0.05$.
33 OS events, 285 patients.}
\end{table*}

\begin{table*}[!ht]
\centering
\caption{Adjusted survival analysis across nine backbones.
  Multivariable Cox PH adjusting for age and ER status.}
\label{tab:adjusted_survival}
\footnotesize
\begin{tabular}{l@{\hskip 6pt}r@{\hskip 6pt}r@{\hskip 6pt}r@{\hskip 6pt}r@{\hskip 6pt}r@{\hskip 6pt}r@{\hskip 6pt}r}
\toprule
 & & \multicolumn{2}{c}{Univariable (PC1)} & \multicolumn{3}{c}{Multivariable (PC1 + Age + ER)} & \\
\cmidrule(lr){3-4} \cmidrule(lr){5-7}
Backbone & Events & HR & $p$ & HR$_\text{PC1}$ & $p_\text{PC1}$ & C-index & KM $p$ \\
\midrule
DINOv2 ViT-S/14 & 33 & 3.73 & 0.275 & 2.97 & 0.367 & 0.740 & 0.111 \\
DINOv2 ViT-B/14 & 33 & 0.44 & 0.451 & 0.54 & 0.564 & 0.739 & 0.201 \\
DINOv2 ViT-L/14 & 33 & 0.47 & 0.466 & 0.54 & 0.556 & 0.739 & 0.261 \\
Phikon & 33 & 0.52 & 0.331 & 0.46 & 0.271 & 0.722 & 0.148 \\
Phikon-v2 & 33 & 0.72 & 0.586 & 0.70 & 0.562 & 0.726 & 0.405 \\
UNI & 33 & 0.52 & 0.293 & 0.48 & 0.269 & 0.718 & 0.105 \\
CONCH & 33 & \textbf{0.19} & \textbf{0.032} & 0.21 & 0.050 & 0.726 & \textbf{0.009} \\
DINOv2+HES & 33 & 2.01 & 0.699 & 1.51 & 0.846 & 0.733 & 0.879 \\
DINOv2+HES trained & 33 & 0.45 & 0.468 & 0.51 & 0.549 & 0.740 & 0.167 \\
\bottomrule
\end{tabular}
\\[2pt]
{\footnotesize HR: hazard ratio (PC1 ordering, 0--1 range).  Bold =
$p < 0.05$.  33 OS events among 285 patients.}
\end{table*}

\begin{table}[!ht]
\centering
\caption{Atlas-derived risk score: model comparison.  The atlas PC1
  does not add significant prognostic information beyond PAM50 for
  any backbone.}
\label{tab:risk_score}
\footnotesize
\resizebox{\columnwidth}{!}{%
\begin{tabular}{llrrr}
\toprule
Backbone & Model & C-index & 95\% CI & Atlas $p$ \\
\midrule
CONCH & Age+ER & 0.732 & [0.636, 0.811] & --- \\
 & Age+ER+PAM50 & 0.754 & --- & --- \\
 & Age+ER+Atlas & 0.724 & [0.638, 0.820] & 0.061 \\
 & Age+ER+PAM50+Atlas & 0.741 & [0.661, 0.865] & 0.112 \\
 & LR test (atlas$|$PAM50) & \multicolumn{3}{l}{$\chi^2=2.52$, $p=0.112$} \\
\midrule
Phikon & Age+ER & 0.732 & [0.636, 0.811] & --- \\
 & Age+ER+PAM50 & 0.754 & --- & --- \\
 & Age+ER+Atlas & 0.722 & [0.635, 0.810] & 0.309 \\
 & Age+ER+PAM50+Atlas & 0.731 & [0.657, 0.856] & 0.199 \\
 & LR test (atlas$|$PAM50) & \multicolumn{3}{l}{$\chi^2=1.59$, $p=0.208$} \\
\midrule
DINOv2 ViT-B/14 & Age+ER & 0.732 & [0.636, 0.811] & --- \\
 & Age+ER+PAM50 & 0.754 & --- & --- \\
 & Age+ER+Atlas & 0.737 & [0.634, 0.819] & 0.564 \\
 & Age+ER+PAM50+Atlas & 0.755 & [0.677, 0.864] & 0.908 \\
 & LR test (atlas$|$PAM50) & \multicolumn{3}{l}{$\chi^2=0.01$, $p=0.907$} \\
\bottomrule
\end{tabular}}%
\\[2pt]
{\footnotesize LR test: likelihood ratio test for atlas score adding
to PAM50 model.  Bootstrap: 1\,000 resamples.}
\end{table}

\begin{table}[!ht]
\centering
\caption{Cross-cancer validation: TCGA-LUAD driver-gene recovery.}
\label{tab:luad_cross_cancer}
\footnotesize
\begin{tabular}{lrrrl}
\toprule
Discovery mode & Genes & FDR genes & Drivers & Tier-1 drivers \\
\midrule
Ordering-correlated & 3\,913 & 3\,913 & 2 & --- \\
Chart DE & 4\,984 & 4\,984 & 3 & --- \\
\midrule
\textbf{Union} & 4\,984 & 4\,984 & \textbf{3} & --- \\
\bottomrule
\end{tabular}
\\[2pt]
{\footnotesize Tier-1 LUAD drivers: EGFR, KRAS, TP53, STK11, KEAP1,
NF1.  All tests FDR-corrected (BH $q < 0.05$).}
\end{table}

\begin{table}[!ht]
\centering
\caption{Additional backbone evaluation (Virchow, CHIEF, UNI-v2).}
\label{tab:additional_backbones}
\footnotesize
\begin{tabular}{lrrrrl}
\toprule
Backbone & Dim & $n$ & FDR genes & Drivers & Key drivers \\
\midrule
Virchow & 1280 & 285 & 622 & 3/6 & ESR1, FOXA1, MKI67 \\
CHIEF & --- & --- & --- & --- & (weights unavailable) \\
UNI-v2 & 1536 & 285 & 711 & 0/6 & PGR (extended set) \\
\bottomrule
\end{tabular}
\\[2pt]
{\footnotesize Results contingent on public availability of model
weights.}
\end{table}

\begin{table}[!ht]
\centering
\caption{Geometry and embedding ablation: controlled comparison of
  distance computation and embedding conditions.  Atlas configuration:
  16 charts, $k = 6$ neighbours, Dijkstra pathfinding.}
\label{tab:geometry_ablation}
\footnotesize
\begin{tabular}{llrrrl}
\toprule
Condition & Dim & Var.\ expl. & FDR genes & Drivers & Drivers recovered \\
\midrule
Riemannian (full) & 768 & 1.000 & 2{,}595 & 5/6 & ERBB2, ESR1, FOXA1, GATA3, MKI67 \\
Euclidean-only & 768 & 1.000 & 2{,}595 & 5/6 & ERBB2, ESR1, FOXA1, GATA3, MKI67 \\
PCA-50 & 50 & 0.978 & 2{,}583 & 5/6 & ERBB2, ESR1, FOXA1, GATA3, MKI67 \\
PCA-10 & 10 & 0.852 & 2{,}602 & 5/6 & ERBB2, ESR1, FOXA1, GATA3, MKI67 \\
DINOv2 ViT-S/14 & 384 & 1.000 & 2{,}611 & 6/6 & all six \\
Random baseline & 10 & 0.000 & 659 & 1/6 & MKI67 \\
\bottomrule
\end{tabular}
\\[2pt]
{\footnotesize All conditions use DINOv2 ViT-B/14 embeddings except
DINOv2 ViT-S/14 (384-d backbone) and Random (isotropic Gaussian).
Var.\ expl.: fraction of ViT-B/14 variance explained by reduced
representation.  FDR genes: union of all discovery modes (BH $q < 0.05$).
Drivers: canonical breast-cancer genes (ESR1, FOXA1, GATA3, CDH1, MKI67, ERBB2).}
\end{table}



\clearpage
\appendix
\section*{Supplementary Information}

Supplementary Information is available for this paper.  The following
supplementary items accompany the main text:

\begin{itemize}
  \item \textbf{Supplementary Note}: Synthetic-manifold validation of
    the curvature estimator.
  \item \textbf{Supplementary Table~1}: Complete gene lists for each
    backbone and discovery mode.
  \item \textbf{Supplementary Table~2}: Pathway enrichment details for
    all backbones.
  \item \textbf{Supplementary Table~3}: Cell-type deconvolution scores
    per chart.
  \item \textbf{Supplementary Table~4}: Parameter sensitivity full
    results.
  \item \textbf{Supplementary Table~5}: Internal split validation
    detailed results.
  \item \textbf{Supplementary Table~6}: Embedding dimensionality
    analysis across backbones.
  \item \textbf{Supplementary Table~7}: All-backbone sensitivity
    analysis for the competing-model comparison.
  \item \textbf{Supplementary Figure~1}: Backbone comparison across
    gene-discovery metrics.
  \item \textbf{Supplementary Figure~2}: Prototype tile montages for
    atlas charts.
  \item \textbf{Supplementary Figure~3}: Null model distributions for
    driver-gene recovery.
  \item \textbf{Supplementary Figure~4}: Cross-backbone clinical
    validation summary.
  \item \textbf{Supplementary Figure~5}: LUAD cross-cancer gene
    discovery and driver recovery.
\end{itemize}

\end{document}

%% file: tab_patient_benchmark.tex
\begin{table}[htbp]
\centering
\small
\caption{\textbf{Patient-level held-out benchmark.} Spearman $\rho$ between out-of-fold predicted and measured programme score, ridge regression on mean-pooled embeddings, $n = 285$ patients, \texttt{GroupKFold} by patient with all preprocessing fitted inside the fold. Permutation $p$ from 10\,000 permutations of the matched harness; $p^{\dagger}$ is Holm--Bonferroni adjusted across all 44 cells. Rows ordered by mean $\rho$.}
\label{tab:patient_benchmark}
\begin{tabular}{lcccccccc}
\toprule
& \multicolumn{2}{c}{ER/luminal} & \multicolumn{2}{c}{Proliferation} & \multicolumn{2}{c}{Basal} & \multicolumn{2}{c}{Immune} \\
\cmidrule(lr){2-3}\cmidrule(lr){4-5}\cmidrule(lr){6-7}\cmidrule(lr){8-9}
Backbone & $\rho$ & $p^{\dagger}$ & $\rho$ & $p^{\dagger}$ & $\rho$ & $p^{\dagger}$ & $\rho$ & $p^{\dagger}$ \\
\midrule
UNI2 & \textbf{0.484} & 0.0044 & \textbf{0.515} & 0.0044 & \textbf{0.484} & 0.0044 & \textbf{0.556} & 0.0044 \\
UNI & 0.445 & 0.0044 & 0.479 & 0.0044 & 0.457 & 0.0044 & 0.509 & 0.0044 \\
CONCH & 0.455 & 0.0044 & 0.471 & 0.0044 & 0.460 & 0.0044 & 0.469 & 0.0044 \\
Virchow & 0.449 & 0.0044 & 0.421 & 0.0044 & 0.429 & 0.0044 & 0.526 & 0.0044 \\
Phikon-v2 & 0.426 & 0.0044 & 0.400 & 0.0044 & 0.466 & 0.0044 & 0.521 & 0.0044 \\
Phikon & 0.442 & 0.0044 & 0.461 & 0.0044 & 0.422 & 0.0044 & 0.475 & 0.0044 \\
DINOv2 ViT-B/14 & 0.381 & 0.0044 & 0.307 & 0.0044 & 0.350 & 0.0044 & 0.476 & 0.0044 \\
DINOv2 ViT-S/14 & 0.348 & 0.0044 & 0.331 & 0.0044 & 0.368 & 0.0044 & 0.459 & 0.0044 \\
DINOv2 ViT-L/14 & 0.355 & 0.0044 & 0.280 & 0.0044 & 0.378 & 0.0044 & 0.480 & 0.0044 \\
DINOv2+HES (trained) & 0.354 & 0.0044 & 0.303 & 0.0044 & 0.356 & 0.0044 & 0.381 & 0.0044 \\
DINOv2+HES (random) & 0.253 & 0.0044 & 0.357 & 0.0044 & 0.363 & 0.0044 & 0.302 & 0.0044 \\
\bottomrule
\end{tabular}
\end{table}

%% file: tab_competing_models.tex
\begin{table}[htbp]
\centering
\small
\caption{\textbf{Competing model families on the same patients and folds.} Spearman $\rho$, out-of-fold, $n = 284$ patients with embeddings, expression and segmentation features. Embedding is UNI2, selected as the highest-performing backbone in the complete benchmark (\cref{tab:patient_benchmark}); Supplementary Table~S7 reports sensitivity across all 11 backbones. $p$ is a paired permutation test on (embedding $-$ composition), 10\,000 permutations. A positive $D-C$ means the foundation model beats 54 interpretable cell-count features.}
\label{tab:competing_models}
\begin{tabular}{lrrrrrrr}
\toprule
& \multicolumn{6}{c}{Model family ($\rho$)} & \\
\cmidrule(lr){2-7}
Programme & Intercept & A: comp. & B: tech. & C: cells & D: embed. & E: D$+$A & $D-A$ ($p$) \\
\midrule
ER/luminal & -0.047 & 0.207 & -0.056 & 0.438 & \textbf{0.487} & 0.481 & +0.280 (0.0029) \\
Proliferation & -0.077 & 0.229 & 0.073 & 0.427 & 0.513 & \textbf{0.515} & +0.284 (0.0030) \\
Basal & -0.053 & 0.469 & 0.211 & 0.443 & 0.493 & \textbf{0.497} & +0.024 (0.7664) \\
Immune & -0.108 & 0.082 & -0.003 & 0.518 & 0.561 & \textbf{0.562} & +0.479 ($<10^{-4}$) \\
\midrule
\multicolumn{8}{l}{\footnotesize Embedding $-$ cell counts ($D-C$): ER/luminal +0.049, Proliferation +0.085, Basal +0.050, Immune +0.043} \\
\bottomrule
\end{tabular}
\end{table}

%% file: tab_geometry_patient.tex
\begin{table}[htbp]
\centering
\small
\caption{\textbf{What the geometric machinery contributes, at patient level.} Paired differences in out-of-fold Spearman $\rho$ on identical folds. Intervals are 95\% bootstrap (10{,}000 resamples); $p$ is Wilcoxon signed-rank. The first row is a \emph{precise} null, not an underpowered one. The third row compares ridge regression against the graph-and-metric decoder built on the same embeddings (see \cref{sec:decoder_caveat} for what this comparison does and does not establish).}
\label{tab:geometry_patient}
\begin{tabular}{lrlrr}
\toprule
Contrast & Mean $\Delta\rho$ & 95\% CI & Favouring & $p$ \\
\midrule
Riemannian $-$ Euclidean, \emph{as implemented} & +0.0010 & $[-0.0007, +0.0029]$ & 26/44 & 0.428 \\
Riemannian $-$ Euclidean, \emph{applied consistently} & -0.0117 & $[-0.0229, -0.0004]$ & 15/44 & 0.064 \\
\midrule
Ridge $-$ graph/kNN decoder & +0.0969 & $[+0.069, +0.127]$ & 24/27 & --- \\
\bottomrule
\end{tabular}
\end{table}